\documentclass[10pt,twocolumn,letterpaper]{article}

\usepackage{cvpr}              % To produce the CAMERA-READY version
\usepackage{graphicx}
\usepackage{amsthm}
\usepackage[accsupp]{axessibility}
\newcommand{\methodName}{OASIS}

\usepackage{microtype}
\usepackage{booktabs, multirow, makecell, xcolor, colortbl, graphicx}

\usepackage{pifont,xcolor}
\newcommand{\cmark}{\textcolor{green!60!black}{\ding{51}}} % ✔
\newcommand{\xmark}{\textcolor{red!70!black}{\ding{55}}}   % ✘

\definecolor{cvprblue}{rgb}{0.21,0.49,0.74}
\usepackage[pagebackref,breaklinks,colorlinks,allcolors=cvprblue]{hyperref}

\def\paperID{5865} % *** Enter the Paper ID here
\def\confName{CVPR}
\def\confYear{2026}

\usepackage{booktabs}
\usepackage{makecell} % 必须加载此包以支持\makecell命令
\usepackage[table]{xcolor} 
\usepackage{adjustbox}
\usepackage{float}
\usepackage{caption}
\usepackage{lipsum}
\usepackage{algorithm}
\usepackage{algpseudocode}
\usepackage{amsmath}

\title{OASIS: On-Demand Hierarchical Event Memory for Streaming Video Reasoning}

\author{
Zhijia Liang$^{1*}$ \quad
Jiaming Li$^{1*}$ \quad
Weikai Chen$^{\ddagger}$ \quad
Yanhao Zhang$^{2}$ \quad
Haonan Lu$^{2}$ \quad
Guanbin Li$^{1,3,4\dagger}$\\
$^1$Sun Yat-sen University\quad $^2$OPPO AI Center \quad $^{3}$Shenzhen Loop Area Institute \\ 
    $^{4}$Guangdong Key Laboratory of Big Data Analysis and Processing \\
 {\tt\small liangzhj56@mail2.sysu.edu.cn,} {\tt\small liguanbin@mail.sysu.edu.cn}
}

\begin{document}
\twocolumn[{%
\renewcommand\twocolumn[1][]{#1}%
\maketitle
\begin{center}
    \captionsetup{type=figure}
    \includegraphics[width=1\linewidth, trim=0 10 0 0, clip]{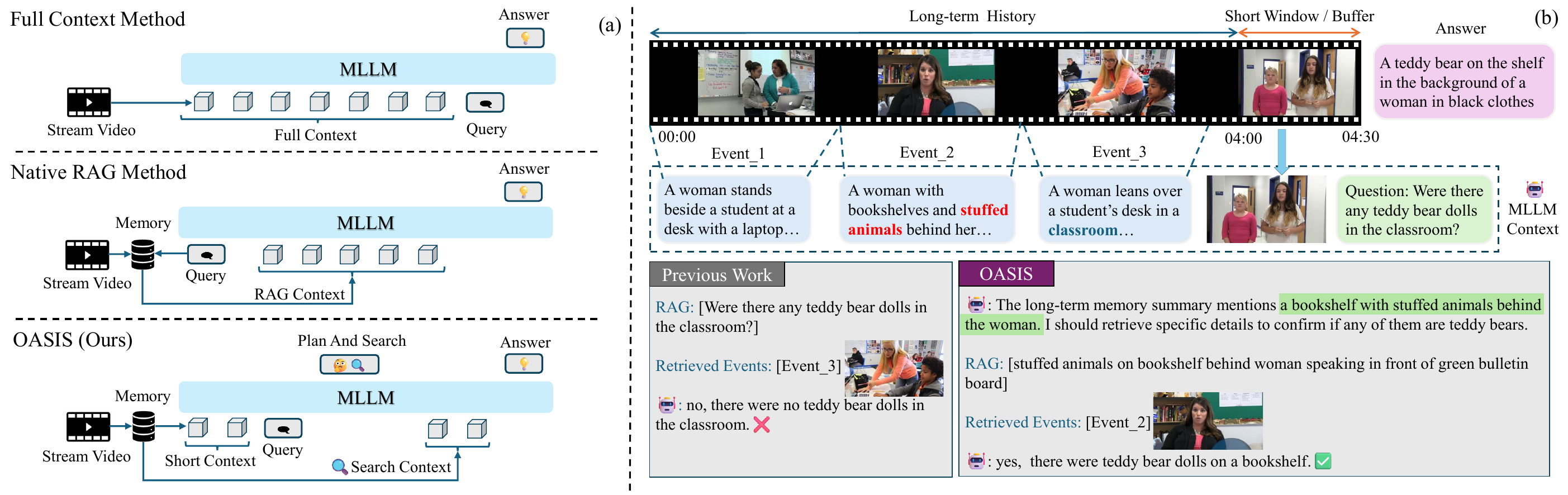}
    \captionof{figure}{(a) Comparison of the \methodName{} framework with conventional streaming video understanding methods. \textbf{Full Context Method:} Blindly stacks all historical frames into the MLLM, leading to high cost and attention collapse. \textbf{Naive RAG Method:} Uses a fixed-policy (e.g., top-k) retrieval that is not task-adaptive. The rigid memory access policy retrieve and stack evidence into a single context window alongside the present-moment frames all the time, inevitably contaminates the reasoning process. \textbf{\methodName{}: }The model first performs a Coarse Reasoning using a Short Context. If insufficient, it initiates a Fine Reasoning phase, infers what additional evidence would make the answer decidable, and uses this semantic hypothesis to route a precise jump into the hierarchical event memory. (b) On the query ``Were there any teddy bear dolls in the classroom?", Naive RAG embeds the raw query and broadly matches ``in classroom", retrieving semantically irrelevant memories. OASIS infer what evidence is needed to solve the problem based on memory, yielding a precise answer.}
\label{fig:teaser}
\end{center}
}]
\renewcommand{\thefootnote}{\fnsymbol{footnote}}
\footnotetext[1]{Equal contribution.}
\footnotetext[2]{Corresponding author is Guanbin Li.}
\footnotetext[3]{This paper solely reflects the author's personal research and is not associated with the author's affiliated institution.}
\renewcommand{\thefootnote}{\arabic{footnote}}

\begin{abstract}
Streaming video reasoning requires models to operate in a setting where history grows without bound while meaningful evidence remains scarce. 
In such a landscape, relevant signal is like an oasis -- small, critical, and easily lost in a desert of redundancy.
Enlarging memory only widens the desert; aggressive compression dries up the oasis. The real difficulty lies in discovering where to look, not how much to remember.
We therefore introduce \methodName{}, a novel framework for streaming video reasoning that tackles this challenge through structured, on-demand retrieval. It organizes streaming history into hierarchical events and performs reasoning as controlled refinement -- short-context inference first, followed by semantically grounded retrieval only when uncertainty arises.
As the retrieval is driven by high-level intent rather than embedding similarity, the retrieved memory is substantially more accurate and less noisy.
% Additionally, the mechanism is plug-and-play, training-free, and compatible with any streaming MLLM. 
% Experiments across multiple benchmarks show that \methodName{} achieves strong gains in long-horizon accuracy and compositional reasoning with far less memory budget. 
Additionally, the mechanism is plug-and-play, training-free, and readily attaches to different streaming MLLM backbones. 
Experiments across multiple benchmarks and backbones show that \methodName{} achieves strong gains in long-horizon accuracy and compositional reasoning with bounded token cost and low request delay. Code is available at \url{https://github.com/Solus-sano/OASIS}.
\end{abstract}    
\section{Introduction}
\label{sec:intro}

Streaming video reasoning lies in the native regime of temporal intelligence -- the world does not present itself in shuffled clips, but as a single continuous stream. 
Autonomous driving, security operators, AR glasses, and embodied agents: all of them need to interpret the world as it unfolds continuously, with no ability to scroll back.
This regime is fundamentally hard as the temporal context grows without bound while the information useful to each query is extremely sparse.
On one hand, na\"{i}vely retaining all past frames causes attention to drown in a desert of noises.
On the other hand, compressing them permanently could erase subtle but decisive evidence.
Simply scaling the token budget or the compression rate only shifts where the failure happens, without resolving the core difficulty: long streaming video is mostly redundant, yet each query usually depends on some tiny and highly localized regions in the past.

% \begin{figure}[t]
% \centering
% % \vspace{-0.25cm}
% % \includegraphics[width=\textwidth, trim=0 0 0 5, clip]{images/pipeline_final.pdf}
% \includegraphics[width=0.92\linewidth, trim=0 10 0 0, clip]{images/teaser_0.pdf}
% % \vspace{-3mm}
% % \vspace{-0.25cm}
% \caption{ 
% Comparison of \methodName{}'s framework with existing streaming video understanding approaches. (a) Full Context Method: Directly stacks all historical video frames into the MLLM. This approach faces severe context length limitations and often leads to attention misalignment for present-moment queries. (b) Naive RAG Method: Retrieves a fixed amount of evidence (e.g., top-k) from memory. This fixed-policy context is not task-adaptive: it retrieves too much unnecessary history for present queries (causing attention misalignment) and is limited by naive retrieval for long-term queries (causing semantic misalignment). (c) \methodName{}: The MLLM first receives an adaptive Short Context (comprising high-resolution present/short-term memory and low-resolution long-term memory summaries) to attempt a direct answer. If insufficient, it initiates a "Plan And Search" phase, generating an intelligent retrieval intent and on-demand retrieving a Search Context. \methodName{}'s two-stage policy effectively resolves the dual dilemmas of "context size" and "context quality".
% }%\vspace{-3mm}
% \label{fig:teaser}
% \end{figure}

Unlike machines, humans do not attempt to retain the totality of past perception. 
Instead, we let almost all of it recede into latent long-term memory and remain focused on the immediate slice of the world that we are currently processing.
Specifically, we operate on a short working window first, and only when we become uncertain do we actively recall a \textit{specific} prior moment -- often a small patch in the timeline that contains the decisive fact. 
In other words, humans do not carry the whole desert; we \textit{find an oasis} when the present demands it. 
Therefore, the critical operation in streaming video reasoning is not storage, compression, or token scheduling, but \textit{on-demand identification} of the correct region in long-term experience. 

This observation motivates a different computational primitive. The essential capability in streaming video reasoning is not to store more history, but to \textit{pinpoint the decisive regions in it}. 
The proper paradigm is thus on-demand and hierarchical event-level retrieval, where the short context serves as the operational substrate, and the long history is accessed only when the present underdetermines the answer.
Viewed this way, streaming video reasoning is best understood as a \textit{temporal routing} problem -- identifying the unique prior event that settles the query.

We materialize this principle by proposing \methodName{} -- \textbf{O}n-Demand \textbf{A}rchitected \textbf{S}emantic \textbf{I}ndexed \textbf{S}lices -- a hierarchical event memory that makes ``oasis seeking" explicit.
\methodName{} maintains three complementary tiers of temporal representation: (i) a high-fidelity \textit{short} window that captures the recent present, (ii) a \textit{medium}-resolution buffer that preserves recent but non-immediate structure, and (iii) a multi-resolution event hierarchy that summarizes \textit{long-term} history.
As the stream unfolds, this hierarchy is updated online via segmentation and structural merging, producing a compact but semantically navigable memory of the past. 

\begin{figure*}[t]
\centering
\includegraphics[width=0.92\linewidth, trim=0 10 0 0, clip]{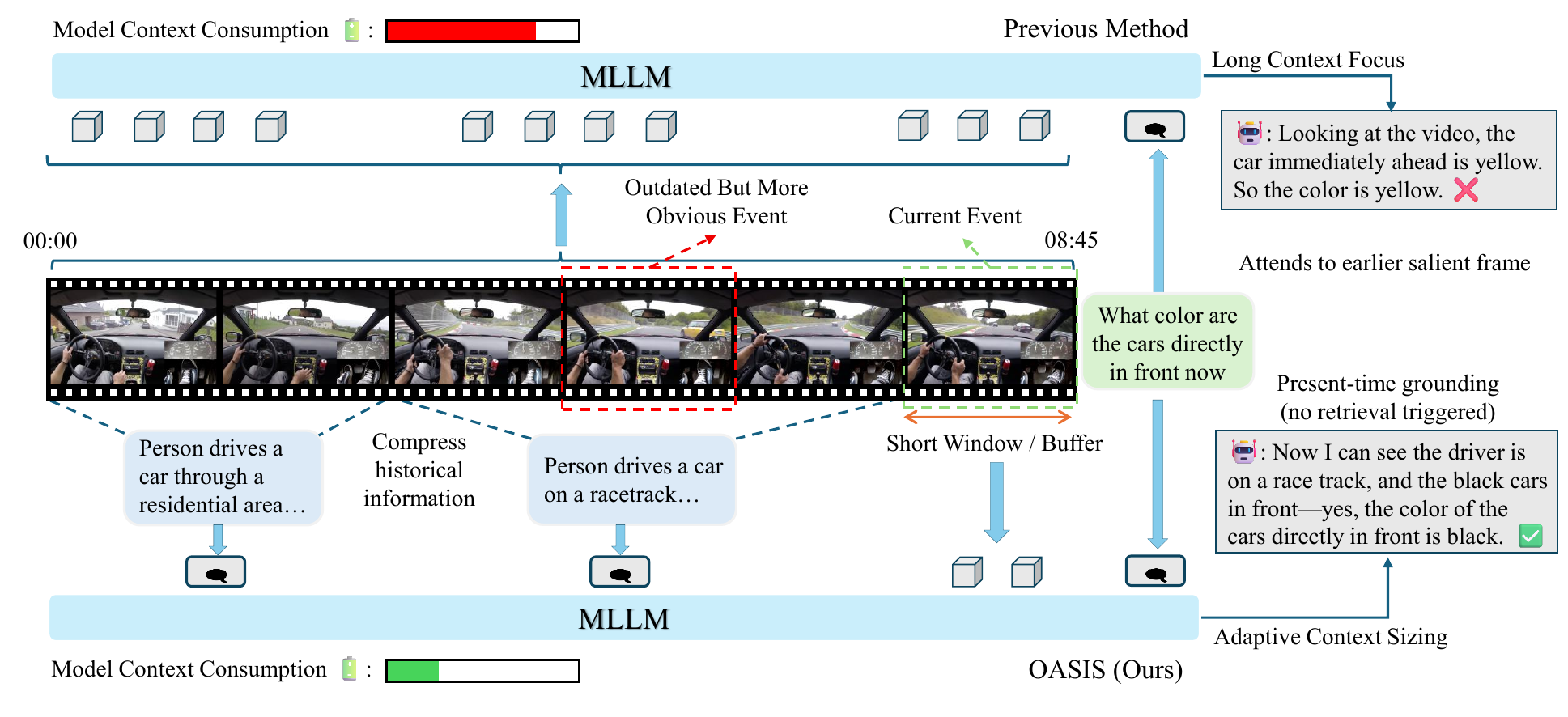}
\caption{ 
A critical failure case of prior methods under real-time querying. When asked, ``\textit{What color are the cars directly in front now?}'', the baseline model (top) processes an excessively large context, but its attention is distracted by a previously salient event (the ``other car''), causing the relevant signal to be drowned in noise and yielding an incorrect answer. In contrast, \methodName{} (bottom) adopts an Adaptive Short Context that constrains attention to the most relevant Short Window/Memory, enabling stable grounding and correct prediction. 
%Demonstrates the critical failure of the previous method on a real-time query. When asked, ``What color are the cars directly in front now
%?", the model (top) consumes massive context, but its attention is distracted by a more ``salient" past event (the ``other car"). This results in attention to drown in a desert of noises and an incorrect answer. In contrast, OASIS (bottom) uses an efficient, Adaptive Short Context, forcing the model to anchor on the ``Short Window/Memory" and answer correctly.
}%\vspace{-3mm}
\label{fig:teaser_2}
\end{figure*}

Critically, \methodName{} executes reasoning in two phases, as shown in Figure~\ref{fig:teaser}(a). The system begins with a short working context, \textit{i.e.} the coarse reasoning stage, where it answers using only the immediately accessible recent window and long-term summary.
As shown in Figure~\ref{fig:teaser_2}, this preserves focus, limits attention noises and attains the fidelity of the present.
When the short context fails to provide sufficient signals, \methodName{} enters a fine reasoning phase. 
Rather than inflating the window wholesale, it infers what additional evidence would make the answer decidable, and uses this semantic hypothesis to route a precise jump into the hierarchical event memory.
The retrieved evidence is thus not obtained through brute-force similarity search, but via semantically guided reasoning that targets the most relevant historical region.
As illustrated in Figure~\ref{fig:teaser}(b), unlike na\"{i}ve RAG-based retrieval that often fetches visually similar but semantically irrelevant memories and yields wrong answers, \methodName{} leverages subtle yet decisive cues to retrieve the correct past event, leading to accurate reasoning over long, cluttered streams.
% \weikai{Here we can refer to Figure~\ref{fig:teaser_2} on comparisons with previous RAG based retrieval.}
This disciplined refinement neutralized both dominate failure modes of streaming video models: attention collapse from unbounded windows, and semantic misalignment from similarity-based retrieval.

% In addition, our formulation is model-agnostic and training-free. \methodName{} does not alter the underlying multimodal large model (MLLM), nor does it require any additional training: it sits on top as a structural memory policy, and can be paired with any streaming MMLM as-is. 
In addition, our formulation is plug-and-play and training-free. \methodName{} does not alter the underlying multimodal large model (MLLM), nor does it require any additional training: it sits on top as a structural memory policy, and can be paired with different streaming MLLM backbones as-is. 
Our empirical results demonstrate that \methodName{} consistently improves long-horizon correctness, lifts reasoning accuracy on compositional queries, and simultaneously reduces the active window budgets by a large margin.
These results validate that structured access is the key determinant of scaling temporal competence.
In summary, our contributions are threefold:
\begin{itemize}
    \item We establish a fundamental reframing of streaming video reasoning as a temporal routing task -- the central difficulty is not capacity, but surfacing the correct region in a vast history.
    \item We introduce \methodName{}, which instantiates this view via a hierarchical event memory and a two-stage refinement protocol that achieves semantically guided, on-demand retrieval.
    % \item We empirically show that \methodName{} is model-agnostic, training-free, and achieves large long-horizon gains across diverse tasks. 
    \item We empirically validate \methodName{} across multiple MLLM backbones, showing that it remains training-free while achieving strong long-horizon gains, bounded token cost, and low request delay. 
\end{itemize}

\section{Related Work}

% \subsection{Multimodal Large Language Models}

% Multimodal Large Models (MLLMs)~\cite{li2023videochat,dai2023instructblip,liu2024kangaroo,zhang2023video,zhang2025videollama,li2024llava,wang2024qwen2} have significantly extended the perceptual capabilities of LLMs by aligning visual encoders with them. For video tasks, due to the enormous number of vision tokens, existing MLLMs typically rely on sparse frame sampling or temporal pooling~\cite{chen2024internvl,zhang2410video,li2024llava,ye2025re,yu2024frame,buch2025flexible,maaz2024video,luo2024video,cheng2024enhancing,wang2025episodic} to stay within restrictive context windows. While effective for global summarization, these designs expose two fundamental limitations. First, aggressive compression or sampling leads to the irreversible loss of fine-grained temporal details. Second, pre-trained primarily on static images or short clips, these models excel at global semantic understanding but lack the dynamic comprehension required for real-time, evolving scenes. Therefore, to use MLLMs as a powerful ``engine" for streaming video, a proper reasoning framework is essential.

\subsection{Long-Video Understanding}

The advent of large language models (LLMs) has catalyzed a paradigm shift across diverse domains~\cite{liguided, yue2025v,vinker2025sketchagent,li2025mccd}. 
In video understanding, MLLMs~\cite{li2023videochat,dai2023instructblip,liu2024kangaroo,
zhang2023video,zhang2025videollama,li2024llava,wang2024qwen2,xie2024large} have made 
significant strides, yet long videos remain challenging due to the large number of visual 
tokens. Existing approaches typically employ sparse frame sampling or 
temporal pooling~\cite{chen2024internvl,zhang2410video,li2024llava,ye2025re,yu2024frame,
buch2025flexible,maaz2024video,luo2024video,cheng2024enhancing,wang2025episodic,wei2023visual} 
to fit within limited context windows. While effective for global summarization, these 
strategies often discard fine-grained temporal evidence critical for downstream reasoning.
Recent works instead explore query-adaptive retrieval over full videos. VideoTree~\cite{wang2025videotree} organizes videos into coarse-to-fine trees to preferentially allocate computation to query-relevant regions. VideoMind~\cite{liu2025videomind} further combines agentic planning with temporal grounding to localize relevant segments in pre-recorded videos. However, both methods assume fixed, fully observed videos and require reprocessing when the stream grows. In contrast, OASIS maintains an online, bounded Event Forest that supports cross-query reuse under continuous streaming without rebuilding search structures.

\subsection{Streaming Video Understanding}

Streaming video understanding requires real-time responsiveness and online interactivity. Early approaches~\cite{chen2024videollm,li2025lion,yang2025svbench,zhao2025cogstream,yao2025timechat} attempt to accommodate all history via stacking or compression, yet escalating token budgets force aggressive summarization, incurring irreversible loss of fine-grained details. Alternative methods~\cite{qian2024streaming,zhang2024flash,zhang2025flash,xiong2025streaming,zheng2025hierarchical} employ external memory with hierarchical structures, but their rigid retrieval-and-stack policy invariably concatenates retrieved cues with current frames, entangling heterogeneous temporal information and inducing attention collapse during reasoning.

% Stacking and compression-based methods~\cite {chen2024videollm,li2025lion,yang2025svbench,zhao2025cogstream,yao2025timechat} attempt to fit all historical information into the MLLM's context. To manage the ballooning token cost, they must employ aggressive compression or sparse sampling. The inherent limitation of this approach is irreversible information loss: fine-grained visual details, once discarded, cannot be recalled for future queries. The second paradigm~\cite{qian2024streaming,zhang2024flash,zhang2025flash,xiong2025streaming} turns to external memory to better organize information. To manage this memory, some advanced works~\cite{zheng2025hierarchical} have even proposed hierarchical memory structures, which is a step in the right direction. However, their primary bottleneck lies not in the memory structure itself, but in their rigid access policy. They are all designed to retrieve and stack evidence into a single context window alongside the present-moment frames. This policy inevitably contaminates the reasoning process, leading directly to the attention collapse. 

% They are still forcing the MLLM to process historical evidence even when answering a simple present-moment query. 
% Consequently, the existing streaming landscape is trapped in this fundamental trade-off. They have failed to find a policy that can decouple present-moment reasoning from long-term retrieval. 

\subsection{Retrieval-Augmented Generation}

Retrieval-Augmented Generation (RAG) enriches LLM responses by retrieving external knowledge and supplying it as conditioning context~\cite{lewis2020retrieval, sarmah2024hybridrag, xie2024weknow, mao2025multi, fan2024videoagent, luo2024video}. However, standard RAG critically assumes that the user’s raw query is already an effective retrieval query—an assumption that breaks down in complex, multi-step interactions. To mitigate this issue, Iterative RAG~\cite{shao2023enhancing} enables the model to propose an initial hypothesis, then repeatedly retrieve new evidence to refine the answer until reaching self-consistency. Inspired by iterative and agentic retrieval paradigms, \methodName{} instantiates intent-driven retrieval for streaming video reasoning with an online-maintained hierarchical memory.
% Building on this paradigm, \methodName{} is the first to introduce this agentic planning concept to the domain of streaming video understanding.  

 % Retrieval-Augmented Generation (RAG)~\cite{lewis2020retrieval,sarmah2024hybridrag,xie2024weknow,mao2025multi,fan2024videoagent,luo2024video} has become a cornerstone for injecting external knowledge into LLMs. However, a key limitation of early RAG approaches is their reliance on the initial user query as the sole basis for retrieval. This assumption is insufficient for video understanding tasks where the optimal evidence is not apparent from the outset. To overcome this, advanced paradigms like Iterative RAG~\cite{shao2023enhancing} have emerged. In this framework, the model adopts an agentic role: it first generates an initial hypothesis or reasoning path, then proactively refines its information needs by retrieving new evidence in subsequent steps. This creates a powerful, self-correcting loop. Inspired by this paradigm, \methodName{} pioneers the adaptation of agentic, iterative retrieval to streaming video understanding. 
\section{Method}
% \chaowei{
%  变量命名原则：

%  - 集合： 大写黑板加粗正体，e.g., $\mathbb X$

%  - 2维以上张量: 大写加粗正体，e.g., $\mathbf X$

%  - 1维向量： 小写加粗正体，e.g., $\mathbf x$

%  - 标量：不加粗斜体，e.g., $x$

%  - 函数：不加粗正体，e.g., $\textrm{ABC}(x)$, $\textrm{F}(x)$

%  - 上标为数字时，添加括号避免和指数运算冲突, e.g., $x^{(1)}$

%  - 关键变量首次出现或未按照以上规则命名时，标明其维度, e.g., $\mathbf X \in \mathbb R^{a\times b\times c}$
%  }
\begin{figure*}[t]
\centering
% \vspace{-0.25cm}
% \includegraphics[width=\textwidth, trim=0 0 0 5, clip]{images/pipeline_final.pdf}
\includegraphics[width=0.95\linewidth, trim=0 0 0 0, clip]{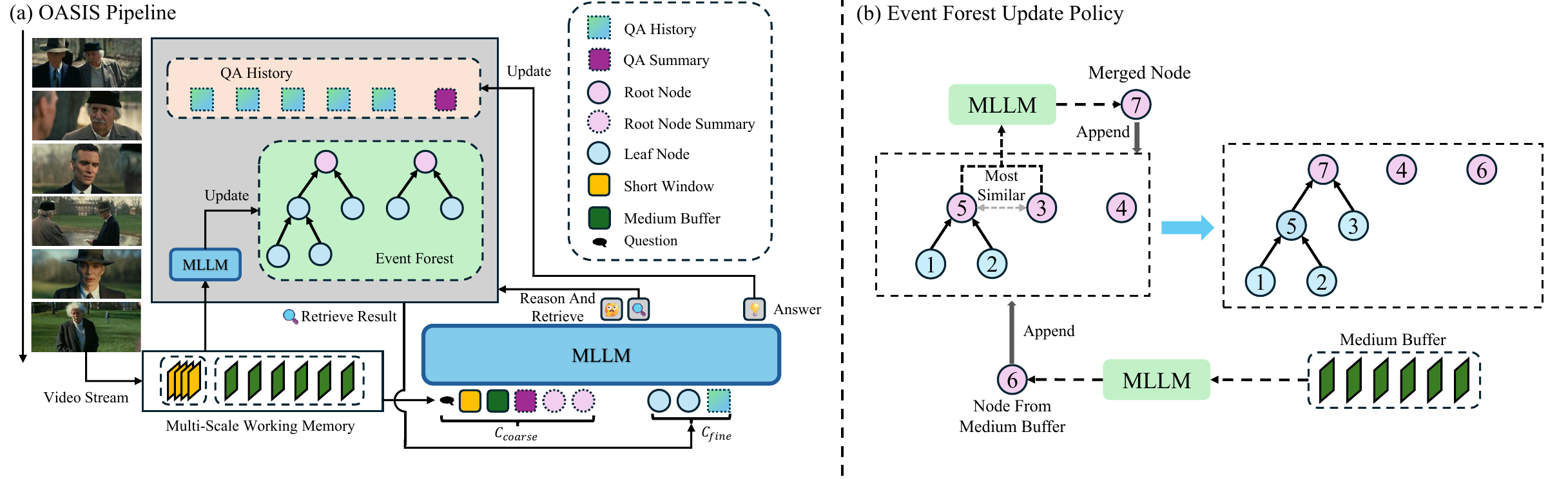}
% \vspace{-3mm}
% \vspace{-0.25cm}
\caption{ 
% The core reasoning and memory maintenance pipeline of \methodName{}. (a) \methodName{} first enters a Coarse Reasoning stage, using the Recent Window and a small set of long-term memory summaries (Event Forest roots). If this is insufficient, it proceeds to the Fine Reasoning stage, retrieving detailed evidence by searching the multi-resolution event hierarchy. (b) To prevent the coarse-stage context from growing too large, OASIS employs a dynamic merging policy. When the number of root nodes exceeds a threshold, the system merges adjacent roots, thereby controlling the size of the root set.
(a) An overview of \methodName{}'s multi-resolution event hierarchy and the two-phase reasoning policy that executes adaptive temporal routing. (b) The dynamic merging strategy used for the online and bounded maintenance of the event forest.
}%\vspace{-3mm}
\label{fig:pipeline}
\end{figure*}

In this paper, we propose \methodName{}, which is a training-free agent system capable of processing streaming video and answering the interleaved questions raised on any moment of the video. Specifically, a hierarchical dynamic memory is introduced to summarize the video stream into a compact yet semantically navigable representation of the past. This memory includes a high-fidelity short window, a medium-resolution buffer, a multi-resolution event hierarchy and a QA summary to capture different resolutions of memory.
Based on the memory, \methodName{} introduces a two-phase reasoning policy which first performs coarse reasoning on the recent window and summative memory and a fine reasoning process to retrieve evidence in the multi-resolution event hierarchy.
The overall framework of \methodName{} is shown in Figure~\ref{fig:pipeline}.

\subsection{Hierarchical Event Memory}
To address the attention collapse problem in multi-modal large language models (MLLM) and achieve efficient long-term memory management, \methodName{} adopts a hierarchical event memory structure, including a high-fidelity short window, a medium-resolution buffer, and a multi-resolution event hierarchy to capture the information in video frames.

\noindent \textbf{Short window. }
% To firmly anchor the agent to the "present," we maintain two frequently updated short-term working memory windows: Current Window ($W$): $W=\{x_t\}_{t=\tau-\ell_w}^{\tau}$.
Given a video $\mathbf{V} = \{\mathbf X_t\}_{t=1}^{n}$ consisting of $n$ video frames(noted as $\mathbf X_t$) and the $i$-th question raised on the $t_i$-th frame,
we introduce a high-fidelity short window to manage the current frames in streaming video. The short window retains the current frames $\mathbf W_s =\{\mathbf X_t\}_{t=t_i - \tau_s}^{t_i}$ on the $t_i$-th frame with a frame rate $r_s$, where $\tau_s$ is a small number indicating the interval of current frames.
The short window preserves the fine-grained visual and temporal details required for accurate, present-moment reasoning.

 \noindent \textbf{Medium-resolution Buffer.}
 Furthermore, \methodName{} introduces a medium-resolution buffer to preserve recent but non-immediate structure. This buffer retains the frames from $ \mathbf W_m =\{\mathbf X_t\}_{t=t_i - \tau_m }^{t_i}$ with a smaller frame rate $r_m$($r_m < r_s$), where $\tau_m$ denotes the interval of medium-resolution frames. In contrast to the short window,  the medium-resolution buffer exploits information from a wider range of frames before the current frame, enlarging the perceptual field of MLLM.

% The short-term window is a medium-duration and medium-frame-rate window that provides richer context about 

% This is an extremely short (e.g., 4-8 seconds) and high-frame-rate window that captures the most immediate visual details to respond to questions about the "here and now." 

% Short-Term Window ($S$): $S=\{x_t\}_{t=\tau-\ell_s}^{\tau}$. 
% This is a medium-duration (e.g., 32 seconds) and medium-frame-rate window that provides richer context for the NowWindow. These two windows constitute the agent's "perceptual field," which is the highest priority input for all subsequent reasoning.

\noindent \textbf{Multi-resolution event hierarchy.}  
% With the short-term memory, the MLLM is capable of performing video understanding. However, the . To address this issue, we propose a long-term memory in an event forest structure.
% A dynamic, structured long-term memory. As video streams continuously flow in, the information in the ShortTermWindow $S$ needs to be compressed and archived into long-term memory. 
% We propose "Event Forest $\mathcal{F}$" as a dynamic, structured memory repository. 
Employing the short-term window and the medium-resolution buffer allow the MLLM to maintain context for recent events. However, for streaming videos, this naive accumulation of information is unsustainable. As the video progresses, the memory required grows linearly, quickly surpassing the context window limitations of modern MLLMs and leading to a significant scalability bottleneck. 
To overcome this, we introduce a multi-resolution event hierarchy, organized as an Event Forest, which summarizes and structures events to maintain a compact yet comprehensive history.

% When the short-term window $\mathbb M_s$ is filled, 
% it is compressed into a level-0 root node $r_i$ and inserted into the root set $\mathcal{R}$ of the forest. 
% Specifically, once the short-term memory $\mathbb M_s$  reaches its predefined capacity, its contents are abstracted into a new event node. This node is initialized as a level-0 root node $r_i$ and subsequently added to the forest's root set $\mathcal{R}$
% Specifically, every $t_n$ frames in $\mathbb{V}$ are abstracted into an event node. 
% Each node $r_i$ is defined as: 

The Event Forest $\mathcal{F}$ consists of a series of event nodes, which are populated on-the-fly by processing the video stream in sequential, non-overlapping temporal windows. Specifically, for every incoming window of $t_m$ frames, we uniform sample $n_f$ key frames and abstract its content into a new event node, which is immediately added to the forest. This ensures that our representation stays synchronized with the live stream. Initially, the $j$-th node $\mathbf R_j$ is constructed by:
\begin{equation}
\mathbf R_j = \big([t^{(j)}_{s},t^{(j)}_{e}], \mathbf{F}_{j}, s_{j}, \mathbf{e}_{j}, d_{j}\big).
\label{eq:node_def}
\end{equation}
Here $[t^{(j)}_{s},t^{(j)}_{e}]$ indicates the time interval of videos abstracted in the node $\mathbf R_j$. Note that $t^{(j)}_{s} - t^{(j)}_{e} = t_m$ initially.
 $\mathbf{F}_{j}$ is a 4D tensor stack of $n_f$ key frames.
 $s_{j}$ is a text summary generated by MLLM that describes the observable facts of the current window. $\mathbf{e}_{j} = \mathbf E(s_j)$ is the embedding vector of this summary output by an embedding encoder $\mathbf E$.  $d_{j}$  denotes the hierarchical level of the node. All nodes created directly from the video stream are initialized as level-0 leaf nodes, i.e., $d_{j} = 0$.  We denote the set of roots as $\mathcal R = \{\mathbf R_j\}_{j=1}^{|\mathcal R|}$.
 
 % $l_{i}$ is the level of the node used for subsequent online merging. The level of a new node generated from the incoming window is set to $l_{i} = 0$.

% \begin{equation}
%   r_i=\big([t^{(i)}_{\text{start}},t^{(i)}_{\text{end}}],~\text{frames}^{(i)},~\text{sum}^{(i)},~\text{emb}^{(i)},l^{(i)}\big).
%   \label{eq:node_def}
% \end{equation}
% Where, $\text{frames}^{(i)}$ is a set of keyframes of size $L_f$ sampled from the window; $\text{sum}^{(i)}$ is a text summary generated by MLLM that only describes the observable facts of the current window; $\text{emb}^{(i)}$ is the embedding vector of this summary; $l^{(i)}$... This indicates the abstraction level of the node; the new node generated by the memory window has a $l$ value of 0.

\noindent \textbf{Node Merging. } 
% To prevent the root set $\mathcal{R}$ from growing indefinitely (which would make the "coarse look" phase too costly), we set a root upper bound $K$. 
% A merge operation is triggered when $|\mathcal{R}|>K$.
% We designed a hierarchy-aware merging strategy: we compute the merging score $s(r_i,r_j)$ for all temporally adjacent root node pairs $(r_i, r_j)$: 
The unbounded growth of the root set would render the context retrieval for the MLLM computationally prohibitive.
To prevent this issue, we introduce a bounded-complexity online merging mechanism to merge the node in the root set.
This mechanism is triggered whenever the number of root nodes 
$|\mathcal{R}|$ exceeds a predefined threshold $n_r$.
We employ a hierarchy-aware strategy that greedily merges the temporally adjacent root pair with the highest merging score. For the temporally adjacent root pair $(\mathbf R_j, \mathbf R_k)$ with $t^{(j)}_{e} = t^{(k)}_{s}$, the merging score is estimated as,
\begin{equation}
    \text{score}(\mathbf R_j,\mathbf R_k) = \cos\big(\mathbf{e}_{j},\mathbf{e}_{k}\big) -\lambda \big(d_{j}+d_{k}\big).
    \label{eq:sim_fun}
\end{equation}
This scoring function elegantly balances two competing objectives. The cosine similarity function $\cos(\cdot)$ promotes the fusion of semantically and temporally coherent events. $\lambda$ is a hyperparameter that penalizes the merging of nodes with high hierarchical levels with large $d_{j}$ and $d_{k}$. This avoids over-abstraction and preserves the hierarchical granularity of memory. 

% The $\cos(\cdot)$ term encourages merging semantically coherent adjacent events. The $\lambda $ penalty term (where $\lambda>0$) inhibits further merging of nodes already at a high level of abstraction (i.e., $l^{(i)}$ or $l^{(j)}$ is large). This avoids over-abstraction and preserves the hierarchical granularity of memory. The system greedily selects the adjacent root pairs with the highest score $s(r_i,r_j)$ to merge, generating a new parent node $r_k$:

The selected pair $(\mathbf R_j, \mathbf R_k)$ is then fused into a new parent node $\mathbf R_l$ which is formulated as,
\begin{align}
    \mathbf R_l = & \big([t^{(l)}_{s},t^{(l)}_{e}], \mathbf{F}_{l}, \mathbf s_{l}, \mathbf{e}_{l},d_{l}\big),  \nonumber\\
   & t^{(l)}_{s} = t^{(j)}_{s},\;\;\;\;t^{(l)}_{e} = t^{(k)}_{e}, \nonumber\\
   & \mathbf{F}_{l} = \mathrm{Uniform}(\mathbf{F}_{j} \oplus \mathbf{F}_{k})\nonumber\\
    &s_{l} = \mathrm{Merge}\big(s_{j},s_{k}\big), \nonumber\\
    &\mathbf e_{l} = \mathbf E(s_{l}), \nonumber\\
    & d_{l} = \max\big(d_{j}, d_{k}\big) + 1.
    % \text{children}^{(k)} &= {r_i, r_j}.
\end{align}
Here $\oplus$ is the concatenation operation. $\mathrm{Uniform}(\cdot)$ is the uniform frame sampling operation. $\mathrm{Merge}(\cdot)$ is the summary merge process performed by LLMs. 
Subsequently, $\mathbf R_j$ and $\mathbf R_k$ are removed from the root set, while the new node $\mathbf R_l$ is added to the root set. The parent-children relation between $\mathbf R_l$ and $(\mathbf R_j, \mathbf R_k)$ is also updated in our event forest $\mathcal{F}$.
This design allows the event forest to adaptively organize long video streams and generate a limited-size summary of the long video by aggregating the summary $\{s_{j}\}_{j=1}^{|\mathcal R|}$ of root nodes.

\noindent \textbf{QA Summary.} We also propose a question-answer(QA) summary to capture the historical question-answer information. For each question $q_i$  raised on the $t_i$-th frame and the corresponding answer $a_i$, OASIS aggregates the content of the question and frame into the current QA Summary by an LLM, 
\begin{equation}
s_{\text{qa}} \leftarrow \mathrm{LLM}\big( s_{\text{qa}}, (q_{i}, a_i)\big).
\end{equation}
The QA summary serves to provide a low-cost dialogue context for MLLM and ensure naming consistency in multi-turn dialogues. Additionally, we also maintain all QA history $\mathcal Q = {\{q_{j}, a_j\}_{j=1}^{i}}$ . Specifically, we estimate an embedding for each question-answer pair by $\mathbf e^{\text{q}}_j = \mathbf E(q_j \oplus a_j)$ for subsequent retrieval.

\subsection{Two-phase Reasoning}
% The core of \methodName{} is its agentic reasoning strategy of ``planning first, then retrieving", which distinguishes it from the naive RAG.
To achieve the ``oasis seeking” explicit in memory context, \methodName{} proposes a two-phase reasoning policy, which operates in two phases. \methodName{} first performs a coarse reasoning on immediately accessible memory and output whether the information is sufficient to answer the question. If so, \methodName{} directly outputs the answer. Otherwise, it will enter the fine Reasoning stage, which retrieves long-term evidence in the multi-resolution event hierarchy to support the answer.

\noindent \textbf{Coarse Reasoning.} 
When a question $q_{i}$ is raised, \methodName{}  applies an MLLM to generate the model response using only the immediately accessible memory. 
% Concretely, the input context $\mathcal{C}_{\text{coarse}}$ for the MLLM consists of all frames in short window and medium-resolution buffer, summaries of all root nodes $\{\mathbf{S}_{i}\}_{i=1}^{n_r}$, and the QA history summary $S_{\text{qa}}$. 
% $Enc$ refers to the visual encoder of $\mathrm{MLLM}$.
Concretely, given the immediately accessible memory $\mathbf{C}_{\text{coarse}}$ (comprising all frames in short window $\mathbf W_s$ and medium-resolution buffer $\mathbf W_m$, summaries of all root nodes $\{s_{j}\}_{j=1}^{|\mathcal R|}$ and the QA history summary $s_{\text{qa}}$) and the raised question $q_{i}$,
the MLLM first outputs a preliminary response with a coarse reasoning process. This process is formulated as,  
\begin{equation} 
Think,~\hat{a}_i \leftarrow \mathrm{MLLM}\big(p_{\text{sys}}, \mathbf{C}_{\text{coarse}}, q_i\big),
\end{equation}
where $p_{\text{sys}}$ is the system prompt. Benefits from our event forest structure, the size of $\{s_{i}\}_{i=1}^{|\mathcal R|}$ is limited, which avoids prohibitive token overhead and attention collapse caused by long input contexts. Subsequently, we modify $p$ to prompt the model to consider two cases.  
If the immediately accessible memory (e.g., current frames or general knowledge) contains all the facts needed to answer $q_{i}$, the model is prompted to directly generate the final answer. 
If the information is insufficient or the question pertains to specific details from the distant past, the model is instructed not to guess. Instead, it must signal the need for a deeper memory search by invoking a specific tool.
The discrimination between these two cases is achieved by parsing special tokens in the model's initial response $\hat{a}$. A response wrapped in $<\text{answer}>$...$<\text{/answer}>$ tags signals that the model has confidently formulated a direct answer, which is then finalized. In contrast, A response containing the $<\text{tool\_call}>$ token indicates that the model has identified the information as insufficient, thereby triggering the fine reasoning phase.
This design forces the model to prioritize high-resolution, recent information, fundamentally suppressing the attentional misalignment of answering the current question with old evidence.

% Otherwise if information is insufficient or the question points to specific details from a distant past, the model is instructed not to guess or be influenced by previous significant events, but must instead enter the ``look closely" phase. 

\noindent \textbf{Fine Reasoning.} 
The fine reasoning is processed in the planning, retrieving, and answering manner.  \methodName{} first generates a retrieval query $I_i$, which can be directly phased from the initial response of the model $\hat{a}_i$. The retrieval query $I_i$ is a task-oriented, high-level semantic internal query, providing more precise retrieval clues than the original question $q_i$.  (e.g., for $q_i$``Was there anyone there just now?" and the video frames of the living room scene, the $I_i$ is set to ``people in the living room"). Subsequently, we encode $I_i$ with the embedding encoder $\mathbf E$ to the embedding $\mathbf E(I_i)$.  This embedding is used to estimate the retrieval score to select the most relevant content in the multi-resolution event hierarchy and the QA history. 

For the multi-resolution event hierarchy, we select the $k_f$ nodes with $I_i$ from the event forest $\mathcal{F}$. To ensure informational diversity and avoid retrieving nodes with direct lineage (e.g., a parent and its child), we employ a greedy pruning strategy.  We iteratively select the node with the highest similarity score, $\text{argmax}_j\cos(\mathbf E(I_i), \mathbf e_j)$, and then remove its ancestors and descendants from the candidate pool for subsequent selections. 
% Let the indices of the resulting $k_f$ hierarchically distinct nodes be $\mathcal I$.
% To ensure informational diversity and avoid redundancy from retrieving nodes with direct lineage (e.g., a parent and its child), we retrieve nodes that are guaranteed to be hierarchically distinct. 
% This is achieved by iteratively selecting the highest-scoring node while simultaneously pruning its ancestors and descendants from the candidate pool for subsequent selections. We denote the indices of the top-$k_f$ node as $\mathbb I$. 
Concurrently, for the QA history $\mathcal Q$,  we retrieve the $k_q$ historical question-answer pairs with top  $k_q$ retrieval scores, $\text{TopK}_j \cos(\mathbf E(I_i), \mathbf e^{\text{q}}_j)$, with $I_i$. 

The retrieval content $\mathbf{C}_{\text{fine}}$ includes the key frames tensor of the $k_f$ selected nodes and the retrived $k_q$ QA pairs. 
Obtained the retrieved content, we update the input context for MLLM and generate the final answer $a_i$,
\begin{equation}
     a_i \leftarrow \mathrm{MLLM}\big(p_{\text{sys}}, \mathbf{C}_{\text{coarse}}, q_i, \mathbf{C}_{\text{fine}}\big).
\end{equation}
This strategic retrieval is the key to achieving accurate and temporally consistent understanding in dynamic video environments.

\section{Experiment}

\begin{table*}[t]
    \centering
    \caption{Comparison on the OVO-Bench benchmark. We report the accuracy(\%) on the Perception and Backward subset. The best results are indicated by \textbf{bolding.} $\ddagger$ denotes results reproduced by us, while others are taken from prior works. }
    \label{tab:ovo-bench}
    \resizebox{\textwidth}{!}{%
    \begin{tabular}{lcccccccccccc}
    \toprule
    \multirow{3}{*}{Model} & \multicolumn{1}{c}{} & \multicolumn{7}{c}{Perception} & \multicolumn{4}{c}{Backward} \\
    \cmidrule(lr){3-9}\cmidrule(lr){10-13}
    & frames & OCR & ACR & ATR & STU & FPD & OJR & Avg & EPM & ASI & HLD & Avg \\
    \midrule
    % Human & - & 93.96 & 92.57 & 94.83 & 92.70 & 91.09 & 94.02 & 93.20 & 92.59 & 93.02 & 91.37 & 92.33 \\
    Gemini 1.5 Pro & 1fps & 85.91 & 66.97 & 79.31 & 58.43 & 63.37 & 61.96 & 69.32 & 58.59 & 76.35 & 52.64 & 62.54 \\
    GPT-4o & 64 & 69.80 & 64.22 & 71.55 & 51.12 & 70.30 & 59.78 & 64.46 & 57.91 & 75.68 & 48.66 & 60.75 \\
    LLaVA-Video-7B  &   64   & 69.13 & 58.72 & 68.83 & 49.44 & \textbf{74.26} & 59.78 & 62.32 & 56.23 & 53.47 & 51.53 & 53.74 \\ 
    LLaVA-OneVision-7B   &   64   & 66.44 & 57.80 & 73.28 & 53.37 & 71.29 & 61.96 & 62.04 & 56.24 & 55.41 & 21.51 & 43.71 \\
    Qwen2-VL-7B  &   64   & 40.40 & 50.46 & 63.76 & 47.19 & 66.34 & 53.55 & 50.98 & 48.71 & 35.45 & 45.03 & 43.73 \\
    Qwen2-VL-72B  &   64   & 65.77 & 60.55 & 69.83 & 51.69 & 69.31 & 54.35 & 61.92 & 52.53 & \textbf{60.81} & \textbf{57.53} & 56.95 \\
    Flash-VStream-7B  &  1fps  & 24.16 & 29.36 & 28.45 & 33.71 & 25.74 & 28.80 & 28.37 & 39.06 & 37.16 &  5.91 & 27.38 \\
    VideoLLM-online-8B &  2fps  &  8.05 &  2.83 & 12.04 & 14.04 & 45.54 & 21.10 & 22.22 & 18.80 & 12.18 & 17.73 & 16.24 \\
    StreamChat$^\ddagger$ & 15fps & 51.68 & 47.71 & 58.62 & 41.01 & 61.06 & 47.83 & 50.12 & 50.17 & 54.05 & 37.63 & 47.39 \\
    Dispider  &  1fps  & 57.72 & 49.54 & 62.07 & 44.94 & 61.39 & 51.63 & 54.55 & 48.48 & 55.41 & 45.43 & 36.06 \\
    \midrule
    Qwen2.5-VL-7B$^\ddagger$  & 0.5fps & 76.51 & 57.80 & 68.10 & 46.63 & 66.34 & 56.52 & 60.93 & 49.83 & 58.11 & 44.62 & 50.24 \\
    \rowcolor{gray!10}
    \textbf{+ \methodName{}} &  -  & 85.23 & 72.48 & 66.38 & 52.25 & 67.33 & 64.67 & 67.26 & 51.85 & 58.78 & 48.92 & 52.61     \\
    
    \midrule
    Qwen3-VL-8B$^\ddagger$    &  0.5fps   & 83.89 & 58.72 & 70.69 & 56.18 & 69.31 & 64.13 & 66.79 & 51.18 & \textbf{60.81} & 43.55 & 51.19 \\
    \rowcolor{gray!10}
    \textbf{+ \methodName{}} &  -  & \textbf{91.95} & \textbf{80.73} & \textbf{81.03} & \textbf{67.42} & 67.33 & \textbf{79.89} & \textbf{78.14} & \textbf{61.95} & 60.14 & 47.31 & \textbf{57.21}     \\
    \bottomrule
    \end{tabular}%
    }
\end{table*}

\begin{table}[t]
    \centering
    \caption{Comparison on StreamingBench. We report accuracy(\%) on All Real-Time Video Understanding (Real-Time All), Misleading Context Understanding (MCU), Anomaly Context Understanding (ACU), and Sequential Question Answering (SQA). $^\dagger$ denotes a dynamic frame rate: 1 fps for durations under 5 min, 0.5 fps for 5--10 min, and 0.2 fps for over 10 min.The best scores are highlighted in \textbf{bold}. $\ddagger$ denotes results reproduced by us, while others are taken from prior works.}
    \label{tab:StreamingBench}
    \setlength{\tabcolsep}{4pt}
    \renewcommand{\arraystretch}{1.05}
    \resizebox{\columnwidth}{!}{
    \begin{tabular}{lccccc}
    \toprule
    Model & frames & Real-Time All & ACU & MCU & SQA \\
    \midrule
    % Human & - &  91.46 & 88.80 & 90.40 & 95.00 \\
    Gemini 1.5 pro   & 1 fps & 75.69 & 51.41 & 40.73 & 54.80 \\
    GPT-4o & 64 & 73.28 & 41.20 & 38.40 & 32.80 \\
    LLaVA-OneVision-7B & 32 & 71.12 & 35.60 & 36.00 & 27.27 \\
    Qwen2-VL-7B & 0.2--1 fps$^\dagger$ & 69.04 & 31.20 & 26.00 & 39.60 \\
    InternVL-V2-8B & 16 & 63.72 & 32.00 & 31.20 & 32.32 \\
    Video-LLama2-7B & 32 & 49.52 & 24.80 & 26.80 & 18.67 \\
    StreamChat$^\ddagger$ & 15 fps & 60.14 & 28.80 & 22.40 & 32.00 \\
    \midrule
    Qwen2.5-VL-7B$^\ddagger$ & 0.5 fps & 65.99 & 33.20 & 30.12 & 34.00 \\
    \rowcolor{gray!10}
    \textbf{+ \methodName{}} & - & 70.59 & 38.80 & 36.00 & 46.00 \\
    \midrule
    Qwen3-VL-8B$^\ddagger$ & 0.5 fps & 72.83 & 35.63 & 35.74 & 43.90 \\
    \rowcolor{gray!10}
    \textbf{+ \methodName{}} & - & \textbf{78.22} & \textbf{42.74} & \textbf{49.60} & \textbf{48.40} \\
    \bottomrule
    \end{tabular}%
    }
\end{table}

\subsection{Benchmark}

To comprehensively evaluate \methodName{}'s ability, we select three streaming video benchmarks: OVO-Bench~\cite{niu2025ovo}, StreamingBench~\cite{lin2024streamingbench}, and StreamBench~\cite{xiong2025streaming}. 
% In all experiments, we strictly adhere to the Online Protocol enforced by these benchmarks: when answering a query $q_i$ at timestamp $t_i$, the model is strictly forbidden from accessing any future video content beyond $t_i$.

% \noindent \textbf{OVO-Bench.} OVO-Bench focuses on time-sensitive online understanding. It categorizes evaluations into Backward Tracing, Real-Time Understanding, and forward active responding. We focus our experiments on the first two passive-response scenarios, as they serve as the perfect touchstone for \methodName{}'s core capabilities: Real-time Understanding directly probes the model's ability to overcome attention misalignment and precisely ground the present moment. Backward tracing directly tests its long-term memory retrieval.
\noindent \textbf{OVO-Bench.} OVO-Bench focuses on time-sensitive online understanding. We focus on the Real-time Understanding and Backward Tracing scenarios.
% as they serve as the perfect touchstone for \methodName{}'s core capabilities
Real-Time Understanding directly probes the model's ability to overcome attention collapse and precisely ground the present moment. Backward Tracing directly tests its long-term memory retrieval.

\noindent  \textbf{StreamingBench.} StreamingBench systematically evaluates streaming understanding across real-time vision, multi-source input, and context comprehension. It simulates continuous multi-turn QA interactions by providing 5 questions per video. We adhere to the official All Context setting, where the entire video stream is made available to the model. 

\noindent \textbf{StreamBench.} We additionally evaluate on StreamBench, a open-ended benchmark; detailed settings and results are deferred to the supplementary material due to space limits.

\subsection{Implementation Details}

% Our experiments are based on the Qwen3-VL-8B~\cite{Qwen3-VL-2025} and Qwen2.5-VL-7B~\cite{bai2025qwen2} models. 
Our experiments are based on the Qwen3-VL-8B~\cite{Qwen3-VL-2025}, Qwen2.5-VL-7B~\cite{bai2025qwen2}, and GLM-4.6V~\cite{hong2025glm}. 
% All MLLM components (Visual Encoder, Projector, and LLM) are directly initialized from official pre-trained weights without any fine-tuning. 
For \methodName{}'s hierarchical memory, we set the short window ($\mathbf W_s$) duration to $\tau_s=8$s at $r_s=2$ fps, and the medium-resolution buffer ($\mathbf W_m$) duration to $\tau_m=32$s at $r_m=1$ fps. Each event node in the Event Forest ($\mathcal{F}$) has a keyframe budget of $n_f=16$, the root set is capped at $n_r=4$ nodes, and the merge penalty is $\lambda=0.1$. To maximize efficiency, the same MLLM instance is used for all management tasks (summarization, merging, QA updates) and final reasoning. We use Qwen3-Embedding-0.6B~\cite{zhang2025qwen3} as the embedding encoder $\mathbf E$. During the Fine Reasoning stage, we retrieve $k_f=2$ event nodes and $k_q=1$ QA pair. All experiments were conducted on a single NVIDIA A800 80GB GPU.
% , accelerated with BFloat16 precision and FlashAttention-2~\cite{dao2023flashattention}.

\subsection{Baselines}
To provide a comprehensive comparison, we benchmark \methodName{} against two categories of state-of-the-art (SOTA) models. First, we compare against SOTA Offline MLLMs to establish a performance ceiling based on general visual reasoning. This includes top-tier open-source models: LongVA~\cite{zhang2024long}, LLaMA-VID~\cite{li2024llama}, LLaVA-Hound~\cite{zhang2025direct}, LLaVA-Video-7B~\cite{zhang2410video}, InternVL2~\cite{chen2024internvl}, LLaVA-OneVision-7B~\cite{li2024llava}, VideoLLaMA2~\cite{zhang2023video}, MiniCPM-V~2.6~\cite{yao2024minicpm}, VILA-1.5~\cite{lin2024vila}, InternLM-XCP2.5~\cite{zhang2024internlm}, MovieChat~\cite{song2024moviechat}, FreeVA~\cite{wu2024freeva}, Qwen2.5-VL~\cite{bai2025qwen2}, Qwen2-VL~\cite{wang2024qwen2}. As well as powerful closed-source models like Gemini-1.5 Pro~\cite{team2024gemini} and GPT-4o~\cite{openai_gpt4o_2024}. Second, we also conduct a head-to-head comparison with SOTA Online MLLMs. This includes Flash-VStream~\cite{zhang2024flash}, VideoLLM-online~\cite{chen2024videollm}, Dispider~\cite{qian2025dispider}, and StreamChat~\cite{xiong2025streaming}. We strictly adhere to the online protocol to simulate online inference: for a query $q_i$ at timestamp $t_i$, the model is only given access to the video segment $\mathbf V_{t_i} = \{\mathbf X_t\}_{t=1}^{t_i}$ and the preceding QA history $\mathcal{Q}_{t_i}$. 

\subsection{Main Results}

\textbf{Results on OVO-Bench. }We compare \methodName{} against existing baselines on OVO-Bench in Table~\ref{tab:ovo-bench}. 
% Our OASIS approach uses an adaptive video stream sampling strategy that combines short windows $\mathbf W_s$ with high frame rates and medium-resolution buffer $\mathbf W_m$ with low frame rates and event forest. Therefore, fps is not a meaningful cost metric. 
When equipped with the \methodName{} framework, both base models (Qwen2.5-VL-7B and Qwen3-VL-8B) achieve improvements in the Perception and Backward scenarios. The most compelling result comes from the Perception scenario, which explicitly tests the model's ability to ground the present moment without being distracted by history—a direct test of the attention collapse dilemma. With \methodName{}, Qwen3-VL-8B achieves a massive +11.35 gain. This result strongly proves that a superior framework can outperform brute-force context \methodName{}'s Coarse Reasoning stage, by forcing a focus on the high-resolution short Window, which fundamentally solves the attention collapse problem. Concurrently, in the Backward scenario, \methodName{} delivers a +6.02 improvement, demonstrating that its Fine Reasoning stage with plan-driven RAG effectively overcomes the irreversible information loss problem. We note that the gains on Qwen2.5-VL-7B are more modest. As \methodName{}'s agentic policy heavily relies on the base LLM's instruction-following capabilities, which are stronger in the Qwen3-VL model.

\begin{table}[t]
    \centering
    \caption{Cross-backbone generalization on OVO-Bench. We report average accuracy(\%) on the Perception and Backward subsets. $\ddagger$ denotes results reproduced by us.}
    \label{tab:glm}
    \setlength{\tabcolsep}{5pt}
    \begin{tabular}{lccc}
    \toprule
    Model & frames & Perception Avg & Backward Avg \\
    \midrule
    GLM-4.6V$^\ddagger$ & 0.5 fps & 54.96 & 51.35 \\
    \rowcolor{gray!10}
    \textbf{+ \methodName{}} & - & \textbf{68.39} & \textbf{55.27} \\
    \bottomrule
    \end{tabular}
\end{table}

\noindent \textbf{Cross-backbone generalization. }To verify that OASIS is not tied to the Qwen family, we further evaluate GLM-4.6V on OVO-Bench in Table~\ref{tab:glm}. OASIS improves GLM-4.6V by +13.43 on Perception and +3.92 on Backward, showing that the gains come from the proposed memory policy rather than a backbone-specific implementation. Together with the Qwen results above, this supports the plug-and-play nature of OASIS across different MLLM families.

\begin{table}[t]
    \centering
    \caption{Ablation study on OVO-Bench. We quantify the impact of each memory component in our \methodName{} system. `Perception Avg' and `Backward Avg' denote the average performance on the Perception and Backward subset, respectively.}
    \label{tab:ablation}
    \setlength{\tabcolsep}{4pt}
    \renewcommand{\arraystretch}{1.05}
    \resizebox{\columnwidth}{!}{
    \begin{tabular}{cccccc}
    \toprule
    %  Planning RAG & Event Forest & Short Window & Perception Avg & Backward Avg \\
    % \midrule
    % \xmark & \xmark & \cmark & 77.18 & 54.68 \\ % w/o RAG、Long-term Memory
    % \xmark & \cmark & \cmark & 76.19 & 55.22 \\ % w/o RAG
    % \cmark & \cmark & \xmark & 70.37 & 56.89 \\ % w/o "Present" Anchor
    % \cmark & \cmark & \cmark & 78.14 & 57.21 \\

    Medium Buffer & Event Forest & Short Window & Perception Avg & Backward Avg \\
    \midrule
    \cmark & \xmark & \xmark & 74.31 & 54.00 \\
    \xmark & \xmark & \cmark & 77.66 & 51.93 \\
    \xmark & \cmark & \cmark & 76.52 & 48.65 \\
    \cmark & \xmark & \cmark & 77.18 & 54.15 \\
    \cmark & \cmark & \xmark & 70.37 & 56.89 \\ 
    \cmark & \cmark & \cmark & 78.14 & 57.21 \\
    \bottomrule
    \end{tabular}%
    }
\end{table}

\noindent \textbf{Results on StreamingBench. } We also evaluate on StreamingBench in Table ~\ref{tab:StreamingBench}. \methodName{} achieves improvements across all four sub-tasks. The gains are particularly pronounced in the MCU task. This task is specifically designed to induce attention collapse by presenting highly similar, continuous footage. \methodName{}'s success here is again attributable to our two-stage policy. Furthermore, the improvements in ACU task demonstrate that \methodName{}'s short window is effective at capturing real-time, transient events. The significant gains in Sequential Question Answering (SQA) prove that \methodName{}'s Plan-driven RAG and QA Summary mechanism are  superior to na\"{i}ve RAG for handling co-reference and long-term dependencies in multi-turn dialogues.

\noindent \textbf{Results on StreamBench. } We also observe consistent gains on the more open-ended StreamBench benchmark; detailed per-subset results and analysis are provided in the supplementary material.
% The gains in Object Search (OS)—the most semantically complex tasks—directly prove the superiority of \methodName{}'s Plan-driven RAG in resolving the attention collapse dilemma.

\begin{table}[t]
    \centering
    \caption{The ablation of key factors in the OASIS on the OVO-Bench dataset. `Native RAG' denotes directly retrieving events with questions. `Flatten Memory' denotes there is no limit to the number of root nodes in the event forest, allowing the root nodes to grow indefinitely.  }
    \label{tab:ablation_different_factor}
    \setlength{\tabcolsep}{4pt}
    \renewcommand{\arraystretch}{1.05}
    \begin{tabular}{lcc}
    \toprule
     Method & Perception Avg & Backward Avg \\
    \midrule
    Flatten Memory & 76.70 & 55.78 \\
    w/o RAG & 76.19 & 55.22 \\
    Na\"{i}ve RAG & 78.01 & 56.13 \\
    OASIS & 78.14 & 57.21 \\ 
    \bottomrule
    \end{tabular}%
\end{table}

\subsection{Ablations}
We conducted a series of rigorous ablation experiments to deconstruct the OASIS framework and quantify the contribution of each core component. We used OVO-Bench as the primary ablation benchmark. All variants used Qwen3-VL-8B as the inference backbone.

% \noindent \textbf{The Anlaysis  of Two-Stage Reasoning. }Our core argument is that streaming understanding must be decoupled into two stages: ``coarse"and ``fine". We validate this in Table~\ref{tab:ablation}. We compare the full OASIS (row 4) with a ``coarse-only" variant (row 2) that removes the RAG retrieval stage. The results show that the accuracy on the Backward task drops from 57.21 to 55.22. This demonstrates that while the root node summary in the ``coarse" stage (row 2) is superior to having no memory at all (row 1, 54.68), it is insufficient to answer questions that rely on fine historical details. This demonstrates the absolute necessity of the “fine reasoning” stage for recalling long-term memory. Next, we removed the high-frame-rate short window $\mathbf W_s$ from the full OASIS (line 3), forcing the model to rely only on the medium-resolution buffer $\mathbf W_m$ and the long-term summary. The accuracy on the Perception task catastrophically plummeted from 78.14 to 70.37. This drop of -7.77 is strong evidence that short window $\mathbf W_s$ is indispensable as a temporal anchor. Without it, the model cannot address attention collapse or reliably answer the present question.

\noindent \textbf{The ablation of the memory system. }We conduct ablation studies in Table \ref{tab:ablation} to quantify the contribution of each component in the memory system. When using only the high-frame-rate short window $\mathbf W_s$, the Perception score reaches 77.66 while the Backward score is only 51.93. When using only the medium buffer $\mathbf W_m$, Perception drops to 74.31 while Backward increases to 54.00. This comparison indicates that longer sampling improves the ability to answer questions that depend on long-term memory, but it can induce an attention collapse that degrades immediate-scene performance. After incorporating the Event Forest, Backward further improves to 57.21, demonstrating the Event Forest’s effectiveness at modeling long-term dependencies.

\noindent \textbf{The ablation of the reasoning policy. }We conduct an ablation study in Table~\ref{tab:ablation_different_factor} to validate the design of our reasoning strategy. When the RAG stage is removed—i.e., the model answers questions solely based on the short window $\mathbf W_s$, medium buffer $\mathbf W_m$, and the root summaries from the Event Forest, the performance on Backward questions drops noticeably. Replacing our Fine Reasoning process with a na\"{i}ve RAG-based retrieval strategy, which directly uses the original question text for retrieval, reduces accuracy on OVO-Bench’s Backward track from 57.21 to 56.13. This demonstrates that generating retrieval cues conditioned on historical summaries allows the model to locate more relevant past events than using the question alone. When we replace the hierarchical Event Forest with a flattened memory structure—removing the root node limit $n_r$ and allowing unbounded root node growth, both Perception and Backward accuracy decline. The primary reason is that excessive root nodes inflate the token count of the coarse summary, making retrieval less efficient and preventing effective parent–child deduplication during hierarchical reasoning. 
% These results confirm that both our planning-based retrieval and hierarchical memory organization are essential for achieving accurate and efficient reasoning over long video streams.

% \noindent \textbf{Token Consumption.} As shown in Figure~\ref{fig:Average_Token_Count}, we compared the average token consumption of OASIS with the full context baseline (uniform sampling at 0.5 fps). The results are compelling: the baseline's token consumption is linearly correlated with the total video duration, reaching as high as 29,517 on StreamingBench (average 7-8 minutes). In contrast, OASIS's token consumption remains relatively stable at around 10k, only one-third of the baseline.This demonstrates that OASIS successfully shifts the computational cost from processing the entire desert to processing only the oasis, and as the main results show, this 3x efficiency improvement is achieved simultaneously with a significant increase in accuracy.

\begin{figure}
    \centering
    \includegraphics[width=1.0\linewidth]{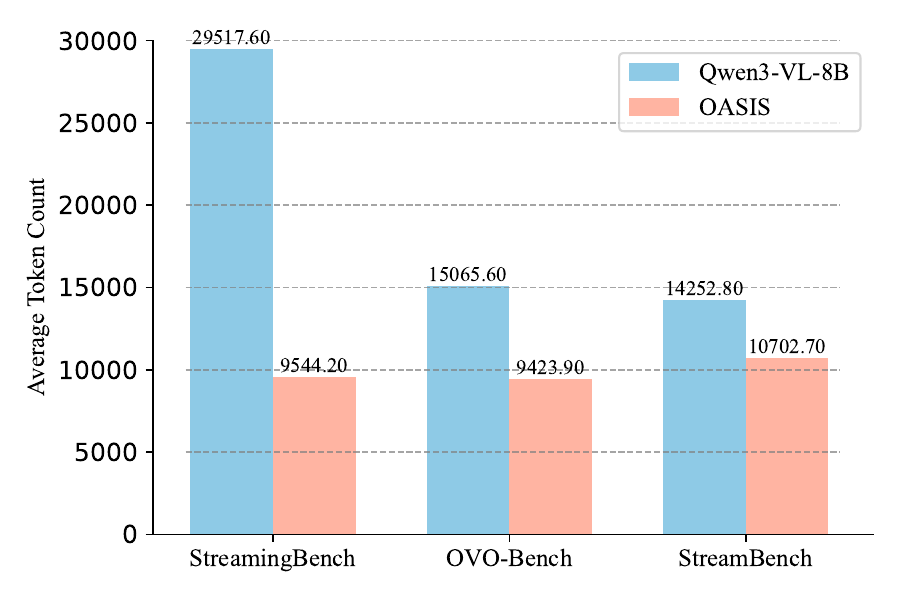}
    \caption{
    Comparison of Average Token Consumption per query between OASIS and the baseline. The Qwen3-VL-8B uniformly samples video at 0.5 fps.
    }
    \label{fig:Average_Token_Count}
\end{figure}

\noindent \textbf{Token Consumption.} As shown in Figure~\ref{fig:Average_Token_Count}, we compare the average token consumption of OASIS against a full-context baseline. The baseline's token usage was positively correlated with video duration, reaching 29,517 tokens on StreamingBench (videos averaging 7–8 minutes) and proportionally high values on OVO-Bench and StreamBench (videos averaging 4–4.5 minutes). In contrast, OASIS's token consumption remains relatively stable at around 10k across all datasets. 
% This behavior demonstrates that OASIS effectively reallocates computation from processing the entire ``desert" to focusing on the informative ``oasis." 

% \noindent \textbf{Token Consumption.} As shown in Figure~\ref{fig:Average_Token_Count}, we compare the average token consumption of OASIS against a full-context baseline (uniform sampling at 0.5 fps). The baseline's token usage was positively correlated with video duration, reaching 29,517 tokens on StreamingBench (videos averaging 7–8 minutes) and proportionally high values on OVO-Bench and StreamBench (videos averaging 4–4.5 minutes). In contrast, OASIS's token consumption remains relatively stable at around 10k across all datasets, roughly one third of the baseline. This behavior demonstrates that OASIS effectively reallocates computation from processing the entire ``desert" to focusing on the informative ``oasis." Notably, this 3× reduction in token cost is achieved concurrently with a substantial improvement in accuracy.

\begin{table}[t]
    \centering
    \caption{Latency and responsiveness of OASIS. RPD is measured on StreamBench following~\cite{xiong2025streaming}; all other latency numbers are measured on OVO-Bench with Qwen3-VL-8B on one A800 GPU.}
    \label{tab:latency}
    \setlength{\tabcolsep}{4pt}
    \scriptsize
    \begin{tabular}{lc}
    \toprule
    Metric & Time (s) \\
    \midrule
    Request Processing Delay (RPD) & \textbf{0.19} \\
    Baseline end-to-end per query & 4.14 \\
    OASIS end-to-end per query & 6.52 \\
    \midrule
    Memory update (generate node) & 7.73 \\
    Memory update (merge root) & 6.52 \\
    Embedding \& RAG retrieval & 0.11 \\
    MLLM inference (coarse) & 3.45 \\
    MLLM inference (fine, if triggered) & 3.06 \\
    \bottomrule
    \end{tabular}
\end{table}

\noindent \textbf{Latency and responsiveness. }Beyond token efficiency, we also analyze wall-clock latency in Table~\ref{tab:latency}. OASIS achieves a low Request Processing Delay (RPD) of 0.19s, indicating that query-time response can start almost once a user question arrives. On OVO-Bench, the end-to-end per-query latency increases from 4.14s for the single-pass baseline to 6.52s for OASIS because of the additional coarse-to-fine reasoning stage. Importantly, the retrieval itself is lightweight, taking only 0.11s; the dominant query-time cost remains MLLM inference (3.45s coarse and 3.06s fine when triggered). The memory-maintenance operations are query-independent and run asynchronously during streaming. With a 32s buffer window, the 7.73s node-generation and 6.52s root-merging costs are overlapped with incoming video.

\section{Conclusion}
\label{sec:conclusion}

In this work, we revisited streaming video reasoning from a structural perspective.
We reformulate this task as temporal routing - the ability to locate the decisive evidence hidden within an overwhelming temporal desert.
Based on this view, we proposed \methodName{}, a hierarchical event memory that enables selective refinement: reasoning  in the short context by default, and retrieving long-term evidence only when semantically required.
The model-agnostic and training-free nature of the approach allows seamless integration with existing MLLMs while delivering substantial efficiency and accuracy improvements.

\section*{Acknowledgments}
This work is supported in part by the National Key
R\&D Program of China (NO.~2024YFB3908503) and in part by the National Natural Science Foundation of China (No.~62322608).

\clearpage
\setcounter{page}{1}
\maketitlesupplementary

% \section{Rationale}
% \label{sec:rationale}
% % 
% Having the supplementary compiled together with the main paper means that:
% % 
% \begin{itemize}
% \item The supplementary can back-reference sections of the main paper, for example, we can refer to \cref{sec:intro};
% \item The main paper can forward reference sub-sections within the supplementary explicitly (e.g. referring to a particular experiment); 
% \item When submitted to arXiv, the supplementary will already included at the end of the paper.
% \end{itemize}
% % 
% To split the supplementary pages from the main paper, you can use \href{https://support.apple.com/en-ca/guide/preview/prvw11793/mac#:~:text=Delete%20a%20page%20from%20a,or%20choose%20Edit%20%3E%20Delete).}{Preview (on macOS)}, \href{https://www.adobe.com/acrobat/how-to/delete-pages-from-pdf.html#:~:text=Choose%20%E2%80%9CTools%E2%80%9D%20%3E%20%E2%80%9COrganize,or%20pages%20from%20the%20file.}{Adobe Acrobat} (on all OSs), as well as \href{https://superuser.com/questions/517986/is-it-possible-to-delete-some-pages-of-a-pdf-document}{command line tools}.

\section{Benchmarks}

\subsection{OVO-Bench}
OVO-Bench~\cite{niu2025ovo} is designed to measure the ability of large video understanding models to comprehend online videos under realistic, temporally constrained conditions. The central protocol is simple: at any arbitrary playback time the model receives a question and must answer using only information available up to that time. OVO-Bench contains 644 independent videos spanning seven major domains (e.g., sports, video games, tutorials). Tasks are organized into three complementary categories, including Backward Tracing, Real-Time Visual Perception, and Forward Active Responding. Backward Tracing probes memory and retrospective reasoning, including Episodic Memory(EPM), Action Sequence Identification(ASI), Hallucination Detection(HLD). Real-Time Visual Perception evaluates the model’s perceptual acuity at the current time and in the immediate past, which focus on Spatial Understanding(STU), Object Recognition(OJR), Attribute Recognition(ATR), Action Recognition(ACR), Optical Character Recognition(OCR), and Future Prediction(FPD). Forward Active Responding assesses a model’s ability to defer response until sufficient evidence accumulates, such as Repetition Event Count(REC), Sequential Steps Recognition(SSR), Clues Reveal Responding(CRR). In this work, we concentrate our experiments on the Backward Tracing and Real-Time Visual Perception categories to study the tradeoffs between episodic recall and instantaneous visual understanding.

\subsection{StreamingBench}
StreamingBench~\cite{lin2024streamingbench} evaluates multimodal large language models (MLLMs) in realistic streaming video, multi-turn interaction settings. The dataset comprises 900 videos and 4,500 questions across eight video types. To emulate continuous interaction, we sample five timestamps per video and pose questions at each. The benchmark is divided into three evaluation axes. Real-Time Visual Understanding measures immediate visual comprehension such as Object Perception(OP), Causal Reasoning(CR), Clips Summarization(CS), Attribute Perception(ATP), Event Understanding(EU), Text-Rich Understanding(TR), Prospective Reasoning(PR), Spatial Understanding(SU), Action Perception(ACP), Counting(CT). Omni-Source Understanding requires synchronous integration of video and audio sources including Emotion Recognition(ER), Scene Understanding(SCU), Source Discrimination(SD), Multimodal Alignment(MA). Contextual Understanding targets robust reasoning under complex temporal contexts, including Misleading Context Understanding (MCU) where earlier frames may introduce distractors, Anomaly Context Understanding (ACU) which requires detecting subtle anomalies, Sequential Question Answering (SQA) where later questions depend tightly on earlier ones, and Proactive Output (PO) which tests when a model should choose to produce an answer. Our experiments emphasize passive Real-Time Visual Understanding and the ACU/MCU/SQA contextual subsets to evaluate model stability in ambiguous or confounding streams.

\subsection{StreamBench}
StreamBench~\cite{xiong2025streaming} focuses on complex, multi-turn interaction in online videos. The collection contains 306 videos and over 1.8k questions, with multiple timestamped queries per video to simulate iterative search and conversational patterns. Questions are open-ended and, for baseline evaluation, we report results using LLaMA3-8B. Tasks fall into six categories: Object Search (OS), Long-term Memory Search (LM), Short-term Memory Search (SM), Conversational Interaction (CI), Knowledge-based QA (KG), and Simple Factual (SF). StreamBench emphasizes the model's ability to maintain retrieval accuracy and dialogue coherence in complex multi-turn interactions with high degrees of freedom.

\section{Long-Horizon Evaluation}

To further validate OASIS beyond the short-to-medium duration benchmarks in the main paper, we additionally evaluate it in two more challenging long-horizon settings. First, we test on HourVideo~\cite{chandrasegaran2024hourvideo}, whose egocentric videos span 20--120 minutes and therefore directly stress temporal reasoning at hour scale. Second, we construct a long-video slice from OVO-Bench by selecting only videos longer than 15 minutes. These two analyses complement the main paper by isolating the regime where long-term memory maintenance and retrieval quality matter most.

\begin{table}[t]
    \centering
    \caption{Results on HourVideo. We report overall accuracy(\%). HourVideo contains egocentric videos spanning 20--120 minutes, providing a direct test of hour-scale temporal reasoning.}
    \label{tab:hourvideo}
    \setlength{\tabcolsep}{10pt}
    \renewcommand{\arraystretch}{1.05}
    \begin{tabular}{lc}
    \toprule
    Model & Accuracy \\
    \midrule
    Qwen3-VL-8B$^\ddagger$ & 35.11 \\
    \rowcolor{gray!10}
    \textbf{+ \methodName{}} & \textbf{37.35} \\
    \bottomrule
    \end{tabular}
\end{table}

\begin{table}[t]
    \centering
    \caption{Results on long clips from OVO-Bench. We select only videos longer than 15 minutes and report accuracy(\%) on the resulting subsets.}
    \label{tab:ovo_long}
    \setlength{\tabcolsep}{5pt}
    \renewcommand{\arraystretch}{1.05}
    \resizebox{\columnwidth}{!}{
    \begin{tabular}{lccc}
    \toprule
    Subset & \#Examples & Qwen3-VL-8B & + \methodName{} \\
    \midrule
    Long-clip perception & 37 & 56.76 & \textbf{70.27} \\
    Long-clip backward & 10 & 30.00 & \textbf{50.00} \\
    \bottomrule
    \end{tabular}
    }
\end{table}

% \noindent \textbf{Hour-scale evaluation.}
% EgoSchema is often associated with long-form video understanding, but each clip is only around three minutes long and therefore does not directly test hour-scale memory. In contrast, HourVideo explicitly targets 20--120 minute videos. 
As shown in Table~\ref{tab:hourvideo}, OASIS improves Qwen3-VL-8B from 35.11\% to 37.35\%, indicating that the proposed hierarchical event memory remains beneficial even when the temporal horizon extends to hour-long videos.

% \noindent \textbf{Long clips from OVO-Bench.}
Table~\ref{tab:ovo_long} further isolates the long-duration regime within a benchmark that otherwise mixes videos of diverse lengths. On videos longer than 15 minutes, OASIS improves the forward/perception subset from 56.76\% to 70.27\% and the backward subset from 30.00\% to 50.00\%. These gains are larger than the average gains reported in the main paper, suggesting that OASIS becomes increasingly advantageous as the amount of historical context grows and naive context accumulation becomes more brittle.

\section{Detailed Results on StreamBench}

StreamBench~\cite{xiong2025streaming} provides a complementary evaluation setting to OVO-Bench and StreamingBench: its questions are open-ended, the task space is more diverse, and the multi-turn interaction pattern is less constrained. This makes it particularly useful for assessing whether OASIS generalizes beyond tightly structured benchmark formats. Following the official protocol, we report semantic-similarity-based evaluation with LLaMA-3-8B as the judge model.

\begin{table*}[t]
    \centering
    \caption{Results on StreamBench. We report accuracy(\%) on 6 subsets. The best scores are highlighted in \textbf{bold}. $\ddagger$ denotes results reproduced by us, while others are taken from prior works. `Avg' denotes the average performance.}
    \label{tab:streambench}
    \resizebox{\textwidth}{!}{%
    \begin{tabular}{lccccccccc}
    \toprule
    Model & frames & OS & LM & SM & CI & KG & SF & Avg \\
    \midrule
    GPT-4o & 50 & 60.5 & 61.2 & 64.4 & 72.3 & 93.9 & 74.7 & 71.0 \\
    LLaMA-VID & 180 & 33.9 & 38.2 & 44.1 & 58.4 & 76.9 & 57.1 & 51.2 \\
    LLaVA-Hound & 8 & 37.6 & 43.2 & 53.4 & 55.7 & 76.3 & 62.0 & 54.7 \\
    LongVA & 8 & 41.1 & 47.4 & 57.6 & 59.8 & 80.7 & 66.1 & 52.4 \\
    MiniCMP-v2.6 & 8 & 37.6 & 51.9 & 43.7 & 65.7 & 66.2 & 64.2 & 56.6 \\
    VILA-1.5 & 8 & 36.1 & 44.4 & 50.8 & 68.3 & 78.6 & 65.5 & 57.1 \\
    InternVL-V2 & 8 & 38.5 & 46.6 & 50.9 & 67.6 & 81.0 & 62.2 & 57.6 \\
    InternLM-XCP2.5 & 8 & 38.8 & 43.3 & 50.8 & 65.6 & 88.4 & 60.5 & 57.7 \\
    MovieChat & 32 & 18.6 & 20.4 & 26.5 & 42.3 & 67.2 & 35.8 & 35.3 \\
    FreeVA & 4 & 35.6 & 37.5 & 43.7 & 58.8 & 84.0 & 53.7 & 56.3 \\
    Video-online & 5 fps & 41.4 & 48.8 & 52.9 & 62.7 & 69.2 & 64.1 & 56.4 \\
    Flash-VStream & 1 fps & 37.1 & 44.5 & 48.6 & 58.1 & 66.4 & 59.2 & 52.1 \\
    StreamChat$^\ddagger$ & 15 fps & 40.7 & 43.6 & 47.1 & 63.0 & \textbf{89.1} & \textbf{73.8} & 59.5  \\
    \midrule
    Qwen2.5-VL-7B$^\ddagger$ & 0.5 fps & 36.1 & 41.3 & 41.0 & 50.5 & 76.5 & 62.3 & 51.1 \\
    \rowcolor{gray!10}
    \textbf{+ \methodName{}} & - & 35.9 & 39.4 & 44.1 & 54.5 & 76.9 & 64.9 & 52.4  \\
    \midrule
    Qwen3-VL-8B$^\ddagger$ & 0.5 fps & 33.8 & 50.5 & 47.8 & 69.4 & 73.5 & 62.6 & 56.0 \\
    \rowcolor{gray!10}
    \textbf{+ \methodName{}} & - & \textbf{47.0} & \textbf{53.6} & \textbf{54.3} & \textbf{71.7} & 86.1 & 67.6 & \textbf{62.1}  \\
    \bottomrule
    \end{tabular}%
    }
\end{table*}

As shown in Table~\ref{tab:streambench}, \methodName{} achieves strong gains on StreamBench across diverse open-ended tasks. In particular, the improvements on Long-term Memory Search (LM) and Short-term Memory Search (SM) show that the coarse-to-fine policy benefits both recent grounding and historical retrieval under freer-form interaction. The gain on Conversational Interaction (CI) further suggests that the QA summary and retrieval mechanism help maintain consistency across multi-turn dialogues. Overall, these results complement the main-paper benchmarks by showing that OASIS remains effective when answer spaces are less constrained and questions require more flexible retrieval and reasoning.

% \begin{table}[t]
%     \centering
%     \caption{The inference latency on OVO-Bench.}
%     \label{tab:latency}
%     \setlength{\tabcolsep}{4pt}
%     \renewcommand{\arraystretch}{1.05}
%     \resizebox{\columnwidth}{!}{
%     \begin{tabular}{cccc}
%     \toprule
%     method & latency(seconds) \\
%     \midrule
%     qwen3-vl-8b & 6.02 \\
%     \methodName{} & 9.55 \\
%     \bottomrule
%     \end{tabular}%
%     }
% \end{table}

\section{Details of Retrieval Algorithm in Fine Reasoning}

\begin{algorithm*}
\caption{Greedy pruning strategy retrieval for Event Forest}
\label{alg:greedy-dedup-retrieval}
\begin{algorithmic}[1]
\Require Event forest $\mathcal{F}$ with node set $\{\mathbf R_j\}$, retrieval query $I_i$, embedding encoder $\mathbf E$, number of retrieved nodes $k_f$
\Ensure Selected event nodes $\mathcal{E}^\star$ (up to $k_f$, hierarchically deduplicated)
\State $\mathcal{C} \leftarrow \{\mathbf R_j \mid \mathbf R_j \in \mathcal{F}\}$ \Comment{initialize candidate set with all nodes}
\State $s_j \leftarrow \cos\!\big(\mathbf E(I_i),\; \mathbf{e}_{j}\big)$ for each $\mathbf R_j \in \mathcal{C}$ \Comment{compute similarity scores}
\State sort $\mathcal{C}$ by $s_j$ in descending order
\State $\mathcal{E}^\star \leftarrow \varnothing$
\While{$|\mathcal{E}^\star| < k_f$ \textbf{and} $\mathcal{C} \neq \varnothing$}
    \State $\mathbf R^\star \leftarrow$ pop highest-scoring node from $\mathcal{C}$
    \State $\mathcal{E}^\star \leftarrow \mathcal{E}^\star \cup \{\mathbf R^\star\}$
    \State $\mathcal{C} \leftarrow \mathcal{C} \setminus \big(\mathrm{Ancestors}(\mathbf R^\star) \cup \mathrm{Descendants}(\mathbf R^\star)\big)$ \Comment{prune lineage nodes}
\EndWhile
\State \Return $\mathcal{E}^\star$
\end{algorithmic}
\end{algorithm*}

To avoid redundancy where a parent node and its children are retrieved simultaneously, we employ a Greedy Pruning Strategy. The algorithm \ref{alg:greedy-dedup-retrieval} illustrates this process: we calculate similarity scores for all nodes in the Event Forest against the retrieval query $I_i$. We iteratively select the highest-scoring node and remove its ancestors and descendants from the candidate pool. This forces the retriever to seek evidence from distinct event branches, ensuring that the final set $\mathcal{E}^\star$ maximizes information diversity while selecting the optimal granularity for each event.

\begin{table}[t]
    \centering
    \caption{The ablation of $\lambda$ on OVO-Bench. Perception Avg and Backward Avg denote the average performance on the Perception and Backward subset, respectively.}
    \label{tab:ablation_lambda}
    \setlength{\tabcolsep}{4pt}
    \renewcommand{\arraystretch}{1.05}
    \begin{tabular}{cccc}
    \toprule
    $\lambda$ & Perception Avg & Backward Avg \\
    \midrule
    0.0 & 78.26 & 57.06 \\
    0.1 & 78.14 & 57.21 \\
    0.2 & 78.74 & 57.37 \\
    0.5 & 78.62 & 57.84 \\
    1.0 & 78.65 & 57.82 \\
    \bottomrule
    \end{tabular}%
    
\end{table}

\begin{table}[t]
    \centering
    \caption{The ablation of $k_f$ on OVO-Bench. Perception Avg and Backward Avg denote the average performance on the Perception and Backward subset, respectively.}
    \label{tab:ablation_kf}
    \setlength{\tabcolsep}{4pt}
    \renewcommand{\arraystretch}{1.05}
    \begin{tabular}{cccc}
    \toprule
    $k_f$ & Perception Avg & Backward Avg \\
    \midrule
    1 & 77.12 & 56.74 \\
    2 & 78.14 & 57.21 \\
    3 & 78.39 & 56.90 \\
    \bottomrule
    \end{tabular}%
    
\end{table}

\begin{table}[t]
    \centering
    \caption{The ablation of $k_q$ on Sequential Question Answering of StreamingBench.}
    \label{tab:ablation_kq}
    \setlength{\tabcolsep}{4pt}
    \renewcommand{\arraystretch}{1.05}
    \begin{tabular}{ccc}
    \toprule
    $k_q$ & SQA Acc \\
    \midrule
    1  & 48.40 \\
    2  & 48.80 \\
    3  & 49.20 \\
    \bottomrule
    \end{tabular}%
    
\end{table}

\section{Analysis of hyper-parameters}
We performed sensitivity analyses on three key hyperparameters $\lambda$, $k_f$, and $k_q$ in the proposed OASIS method to evaluate the robustness of model performance to parameter selection.

\subsection{Sensitivity Analysis of $\lambda$}
$\lambda$ is a hyperparameter that controls the trade-off between the similarity of the event nodes and the hierarchical level of the nodes in the event forest. As shown in Table~\ref{tab:ablation_lambda}, we compare the performance of our method on OVO-Bench under different $\lambda$ values. The performance demonstrates strong robustness to the selection of $\lambda$; across the tested range, the overall difference in Perception Avg is less than $1\%$. 

\subsection{Sensitivity Analysis of $k_f$}
$k_f$ is a hyperparameter that controls the number of event nodes retrieved from the event forest. Table~\ref{tab:ablation_kf} shows the performance variation with respect to $k_f$ on OVO-Bench. Increasing $k_f$ from 1 to 2 yields a significant performance gain, indicating that retrieving more nodes is crucial. However, further increasing $k_f$ to 3 only results in marginal improvement on the Perception subset while slightly decreasing performance on the Backward subset. Thus, we select $k_f = 2$ to maintain an optimal balance between performance and computational efficiency.

\subsection{Sensitivity Analysis of $k_q$}
$k_q$ is a hyperparameter that controls the number of previous questions retrieved from the question memory. As shown in Table~\ref{tab:ablation_kq}, we evaluate the performance of our method on the Sequential Question Answering (SQA) subset of StreamingBench. The results clearly indicate that the SQA accuracy scales positively with the value of $k_q$ within the tested range, achieving the best result of 49.20 at $k_q = 3$.

\section{Computational Efficiency Analysis}
We provide a comprehensive analysis of the computational efficiency of OASIS, focusing on two critical metrics: Peak GPU Memory Footprint and Request Processing Delay~\cite{xiong2025streaming}.
\begin{table}[t]
    \centering
    \caption{\textbf{Component-wise breakdown of Peak GPU Memory usage on OVO-Bench.} We compare the full-context baseline against OASIS with various memory components enabled. Event Forest, Medium Buffer, and Short Window denote the three tiers of our hierarchical memory. The baseline nearly exhausts the A800 memory, while OASIS significantly reduces the footprint.}
    \label{tab:ablation_GPU_Mem}
    \setlength{\tabcolsep}{4pt}
    \renewcommand{\arraystretch}{1.05}
    \resizebox{\columnwidth}{!}{
    \begin{tabular}{ccccc}
    \toprule

    Method & Medium Buffer & Event Forest & Short Window & GPU Memory(GB) \\
    \midrule
    \methodName{} & \cmark & \xmark & \xmark & 20.91 \\
    \methodName{} &\xmark & \xmark & \cmark & 20.09 \\
    \methodName{} &\xmark & \cmark & \cmark & 24.40 \\
    \methodName{} &\cmark & \xmark & \cmark & 22.28 \\
    \methodName{} &\cmark & \cmark & \xmark & 26.15 \\ 
    \methodName{} &\cmark & \cmark & \cmark & 28.48 \\
    Qwen3-VL-8B & - & - & - & 76.59 \\
    \bottomrule
    \end{tabular}%
    }
\end{table}

\subsection{Peak GPU Memory Footprint}
The high memory consumption of long-context MLLMs is the primary bottleneck preventing their deployment on edge devices or standard servers. We measured the peak GPU memory usage during inference on OVO-Bench.

As shown in Table \ref{tab:ablation_GPU_Mem}, the baseline method  consumes a massive 76.59 GB of memory, pushing the limit of a single A800 (80GB) GPU. This excessive footprint is largely due to the linearly growing KV-cache required to store the entire video history.

In contrast, the full OASIS framework consumes only 28.48 GB, a significant reduction in memory requirements. Even with all memory components enabled, OASIS fits comfortably within consumer-grade hardware limits or allows for larger batch sizes on enterprise GPUs. The ablation rows further demonstrate that our memory components are lightweight, with the Event Forest adding minimal overhead compared to the base model weights.

\section{Qualitative Analysis}
We provide concrete visualization examples to demonstrate how OASIS adaptively handles different types of queries in streaming scenarios. We select two representative cases: one requiring Real-time Perception and one requiring Long-term Retrieval.

\begin{figure*}[t]
\centering
% \vspace{-0.25cm}
% \includegraphics[width=\textwidth, trim=0 0 0 5, clip]{images/pipeline_final.pdf}
\includegraphics[width=0.95\linewidth, trim=0 0 0 0, clip]{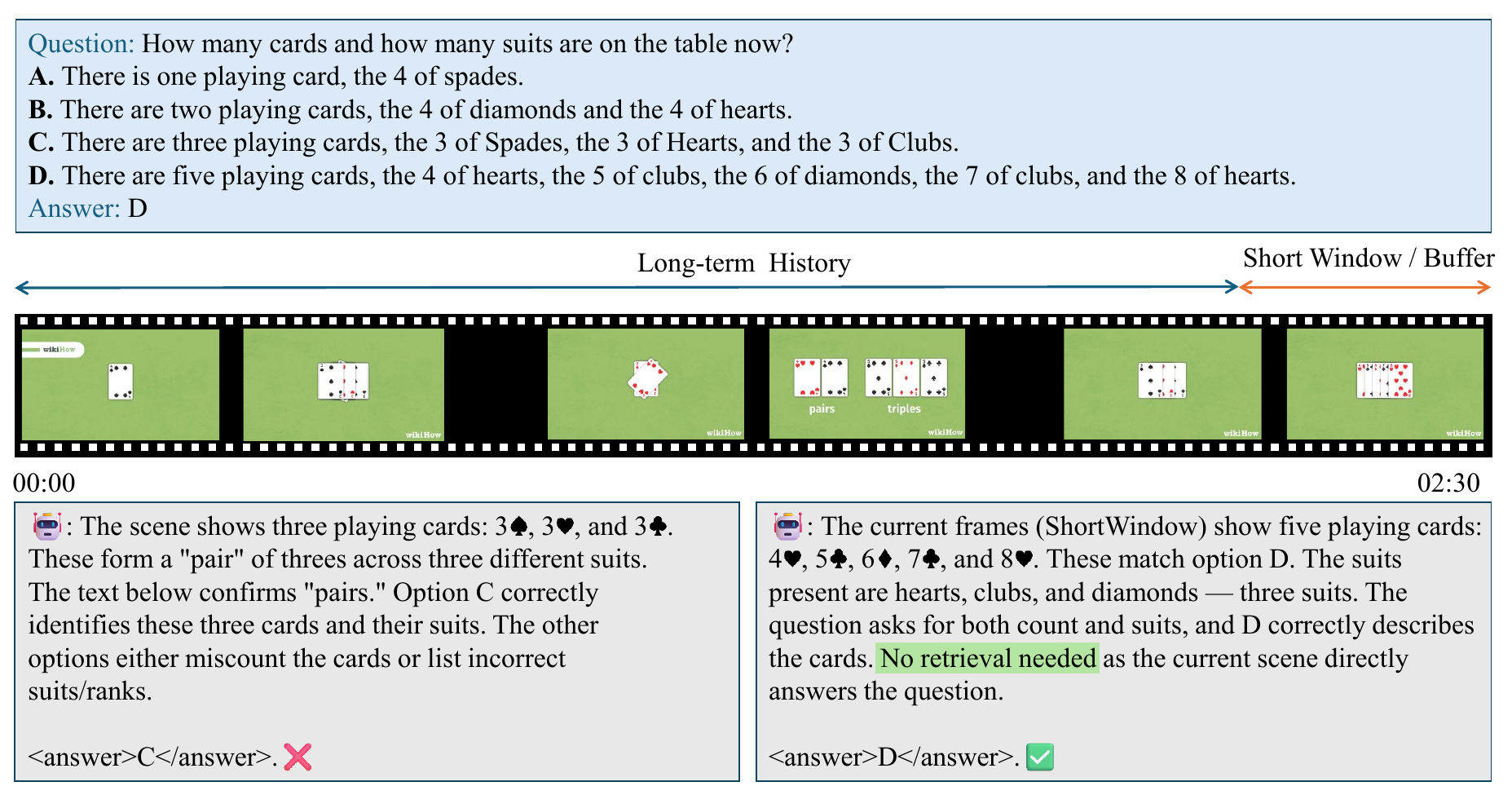}
% \vspace{-3mm}
% \vspace{-0.25cm}
\caption{ 
Visualization of OASIS handling a Real-time Perception query.
}%\vspace{-3mm}
\label{fig:case_0}
\end{figure*}

Figure \ref{fig:case_0} visualizes OASIS handling a real-time perception query where full-context modeling fails due to attention collapse. Despite the explicit temporal cue ``now,'' the baseline misattends to historically-presented cards, yielding an incorrect answer. This failure case underscores the efficacy of our Coarse Reasoning stage: OASIS grounds its reasoning in the Short Window, accurately perceiving the five cards currently on the table. Crucially, the model autonomously assesses this local context as sufficient, thereby avoiding noisy historical information and producing a precise, hallucination-free response.

\begin{figure*}[t]
\centering
% \vspace{-0.25cm}
% \includegraphics[width=\textwidth, trim=0 0 0 5, clip]{images/pipeline_final.pdf}
\includegraphics[width=0.95\linewidth, trim=0 0 0 0, clip]{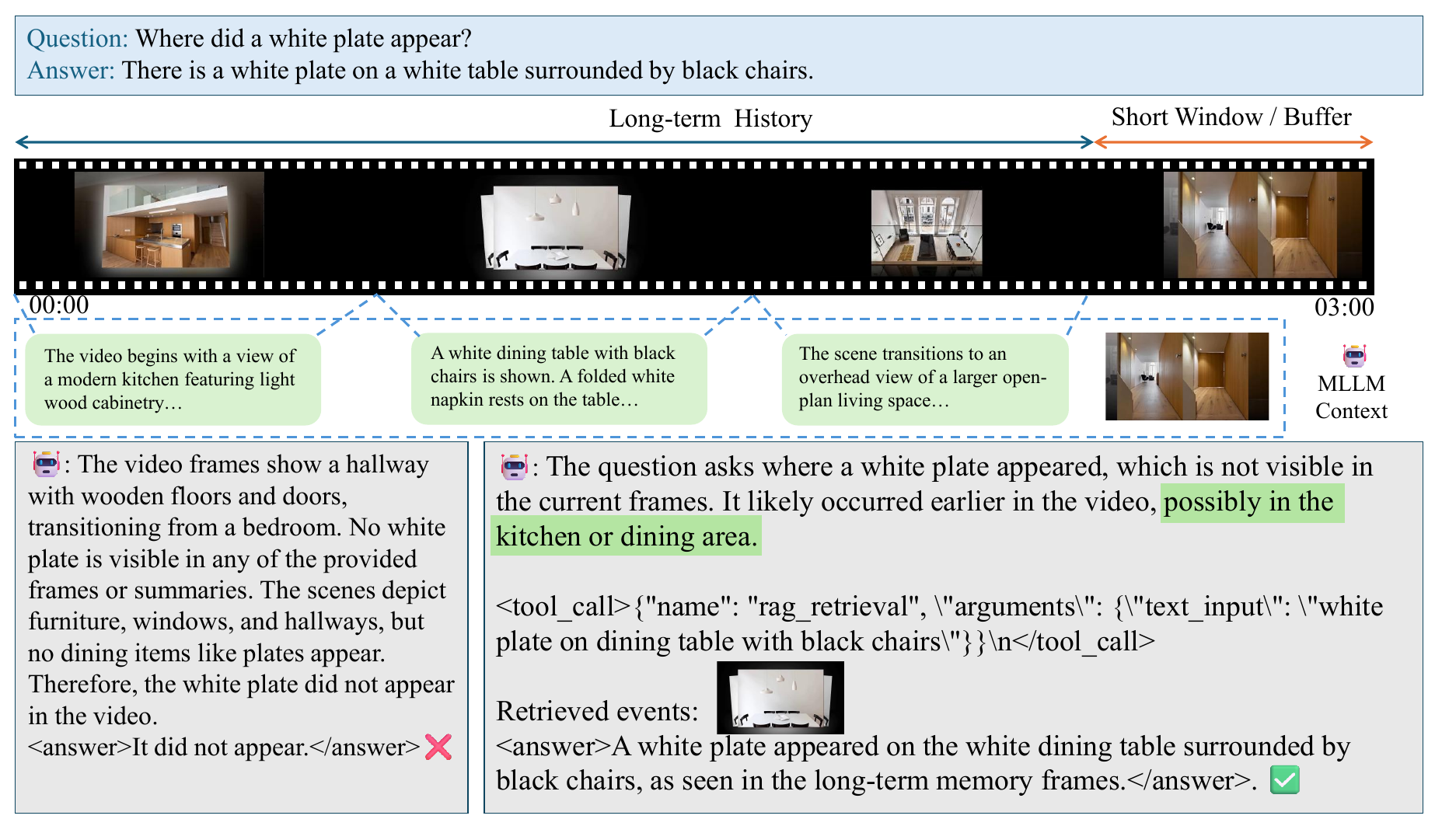}
% \vspace{-3mm}
% \vspace{-0.25cm}
\caption{ 
Visualization of OASIS handling a Long-term Memory query.
}%\vspace{-3mm}
\label{fig:case_1}
\end{figure*}

Figure \ref{fig:case_1} compares the full OASIS framework against its Coarse-reasoning-only variant (left) to demonstrate the necessity of the fine reasoning stage. Without the retrieval capability, the model relies solely on the ShortWindow and summaries, failing due to Evidence Missing and incorrectly concluding the white plate never appeared. In contrast, the full OASIS framework (right) successfully bridges this gap. Upon identifying the information deficiency in the current context, it actively plans a semantic hypothesis, inferring the object likely appeared in a kitchen or dining area, and executes a precise retrieval. This action retrieves the specific dining table event from the long-term history, enabling the model to accurately ground the answer.

\section{Details of Prompts}

\begin{figure*}[t]
\centering
% \vspace{-0.25cm}
% \includegraphics[width=\textwidth, trim=0 0 0 5, clip]{images/pipeline_final.pdf}
\includegraphics[width=0.95\linewidth, trim=0 0 0 0, clip]{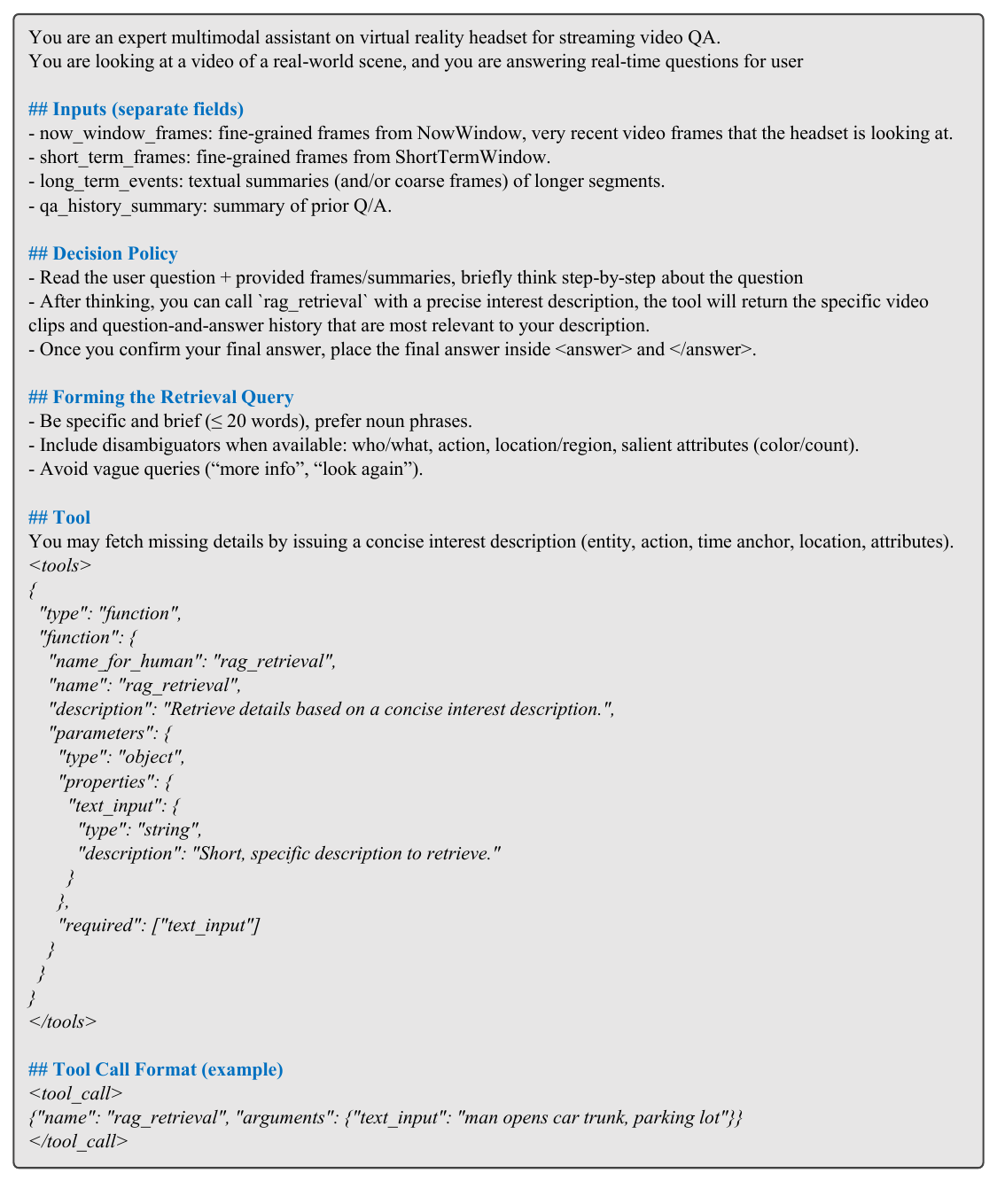}
% \vspace{-3mm}
% \vspace{-0.25cm}
\caption{ 
 System Prompt Template used in OASIS.
}%\vspace{-3mm}
\label{fig:system_prompt}
\end{figure*}

\begin{figure*}[t]
\centering
% \vspace{-0.25cm}
% \includegraphics[width=\textwidth, trim=0 0 0 5, clip]{images/pipeline_final.pdf}
\includegraphics[width=0.95\linewidth, trim=0 0 0 0, clip]{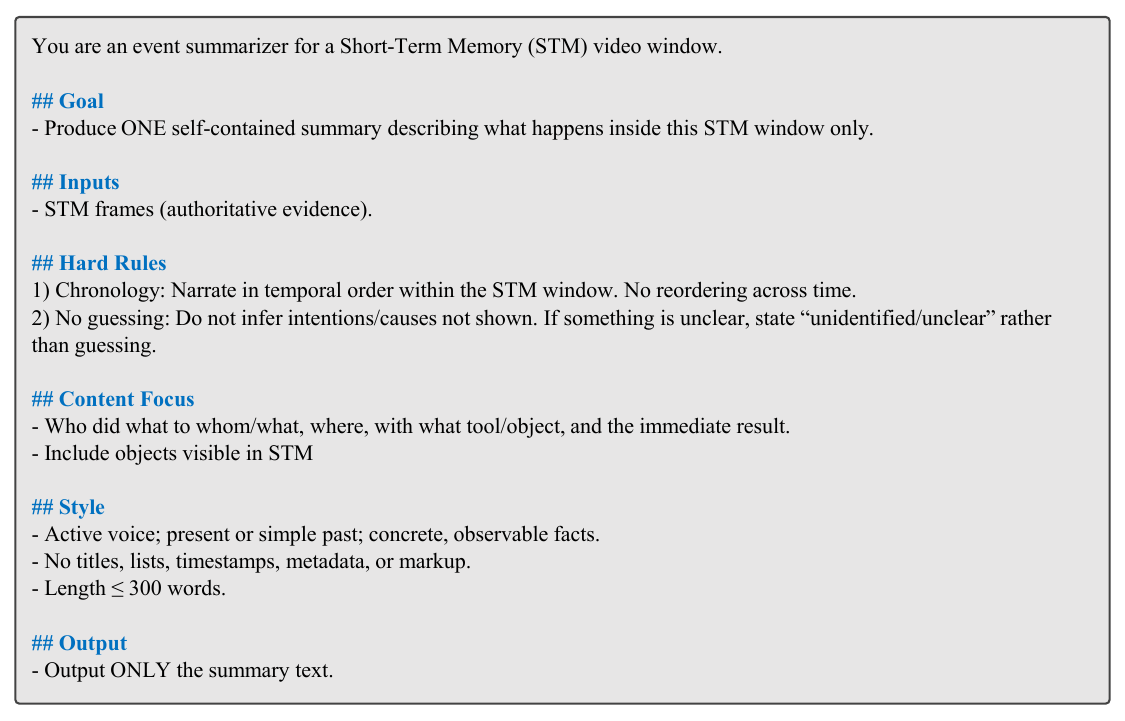}
% \vspace{-3mm}
% \vspace{-0.25cm}
\caption{ 
Event Summary Prompt Template used in \methodName{}.
}%\vspace{-3mm}
\label{fig:prompt_summary}
\end{figure*}

\begin{figure*}[t]
\centering
% \vspace{-0.25cm}
% \includegraphics[width=\textwidth, trim=0 0 0 5, clip]{images/pipeline_final.pdf}
\includegraphics[width=0.95\linewidth, trim=0 0 0 0, clip]{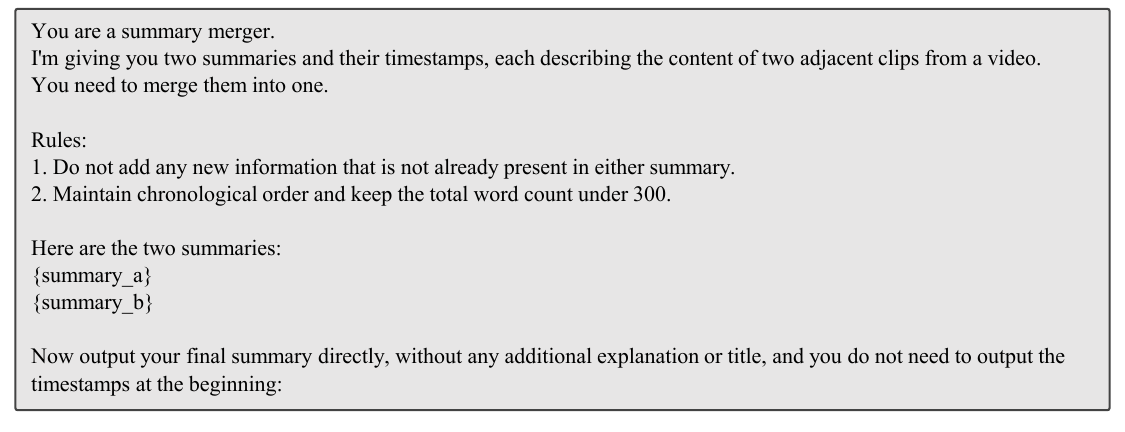}
% \vspace{-3mm}
% \vspace{-0.25cm}
\caption{ 
Event Merge Prompt Template used in \methodName{}.
}%\vspace{-3mm}
\label{fig:prompt_merge}
\end{figure*}

\begin{figure*}[t]
\centering
% \vspace{-0.25cm}
% \includegraphics[width=\textwidth, trim=0 0 0 5, clip]{images/pipeline_final.pdf}
\includegraphics[width=0.95\linewidth, trim=0 0 0 0, clip]{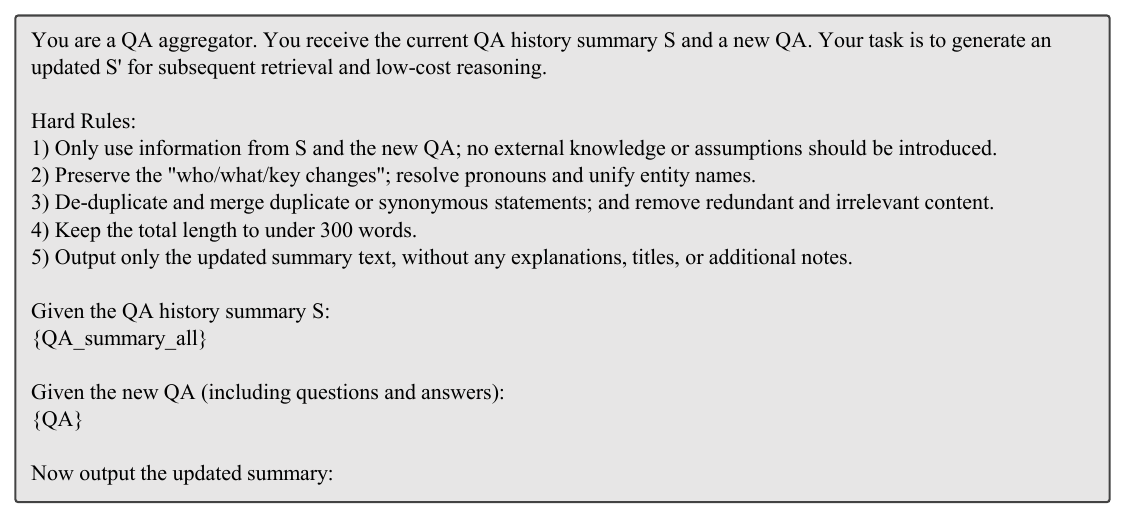}
% \vspace{-3mm}
% \vspace{-0.25cm}
\caption{ 
QA Summary Prompt Template used in \methodName{}.
}%\vspace{-3mm}
\label{fig:prompt_QA}
\end{figure*}

We provide the exact prompts used in our implementation. Figure \ref{fig:system_prompt} illustrates the system instructions for the main inference model, specifically detailing the guidelines for the Two-Stage Reasoning policy. Regarding memory maintenance, Figure \ref{fig:prompt_summary} presents the prompt responsible for generating event nodes from the medium buffer, while Figure \ref{fig:prompt_merge} depicts the instruction for dynamically merging two adjacent event nodes into a unified summary. Figure \ref{fig:prompt_QA} outlines the prompt used for iteratively updating the QA summary.

\section{Future Work}
\methodName{} currently operates in a training-free regime, leveraging MLLM zero-shot reasoning. While effective for standard scenarios, this paradigm exhibits limitations on tasks requiring multi-hop retrieval or fine-grained visual grounding. We will address this by curating instruction-tuning datasets for temporal planning, enabling the model to learn optimal search strategies rather than relying exclusively on prompting. This supervised refinement will enhance the robustness and precision of agentic retrieval in long-form video streams.

{
    \small
    \bibliographystyle{ieeenat_fullname}
    \bibliography{main}
}


\end{document}